\documentclass{article}

\PassOptionsToPackage{sort&compress}{natbib}
\usepackage[preprint]{corl_2026} 
\hypersetup{pdftitle={Shape Your Body: Value Gradients for Multi-Embodiment Robot Design}, pdfauthor={Nico Bohlinger, Jan Peters}}

\usepackage{graphicx}
\usepackage{amsmath}
\usepackage{amssymb}
\usepackage{algorithm}
\usepackage{algpseudocode}
\usepackage{glossaries}
\usepackage{booktabs}
\usepackage{longtable}
\usepackage{multirow}
\usepackage{bbm}

\definecolor{mylightblue}{RGB}{94,165,218}

\glsdisablehyper

\newacronym{rl}{RL}{Reinforcement Learning}
\newacronym{urma}{URMA}{Unified Robot Morphology Architecture}
\newacronym{gnn}{GNN}{Graph Neural Network}
\newacronym{ppo}{PPO}{Proximal Policy Optimization}
\newacronym{sac}{SAC}{Soft Actor-Critic}
\newacronym{pso}{PSO}{Particle Swarm Optimization}
\newacronym{feacrl}{FEACRL}{Fast Evolutionary Actor-Critic Reinforcement Learning}
\newacronym{bo}{BO}{Bayesian Optimization}
\newacronym{cmaes}{CMA-ES}{Covariance Matrix Adaptation Evolution Strategy}
\newacronym{cem}{CEM}{Cross-Entropy Method}
\newacronym{de}{DE}{Differential Evolution}
\newacronym{ars}{ARS}{Augmented Random Search}
\newacronym{turbo}{TuRBO}{Trust Region Bayesian Optimization}
\newacronym{gcpfo}{GC-PFO}{Gradient-Covariance Particle Filter Optimization}
\newacronym{vgds}{VGDS}{Value-Gradient Design Search}
\newacronym{mjx}{MJX}{MuJoCo XLA}
\newacronym{rms}{RMS}{Root Mean Square}

\title{Shape Your Body: Value Gradients for Multi-Embodiment Robot Design}

\author{
  Nico Bohlinger$^{1}$, 
  Jan Peters$^{1,2}$\\
  $^{1}$ Technical University of Darmstadt, Germany\\
  $^{2}$ Robotics Institute Germany (RIG); German Research Center for AI (DFKI); hessian.AI
}

\begin{document}
\maketitle


\vspace{-2.5em}
\begin{center}
\href{https://nico-bohlinger.github.io/shape-your-body/}{\textbf{\textcolor{mylightblue}{nico-bohlinger.github.io/shape-your-body}}}
\end{center}
\vspace{-0.5em} 

\begin{figure}[!h]
\centering
\includegraphics[width=\textwidth]{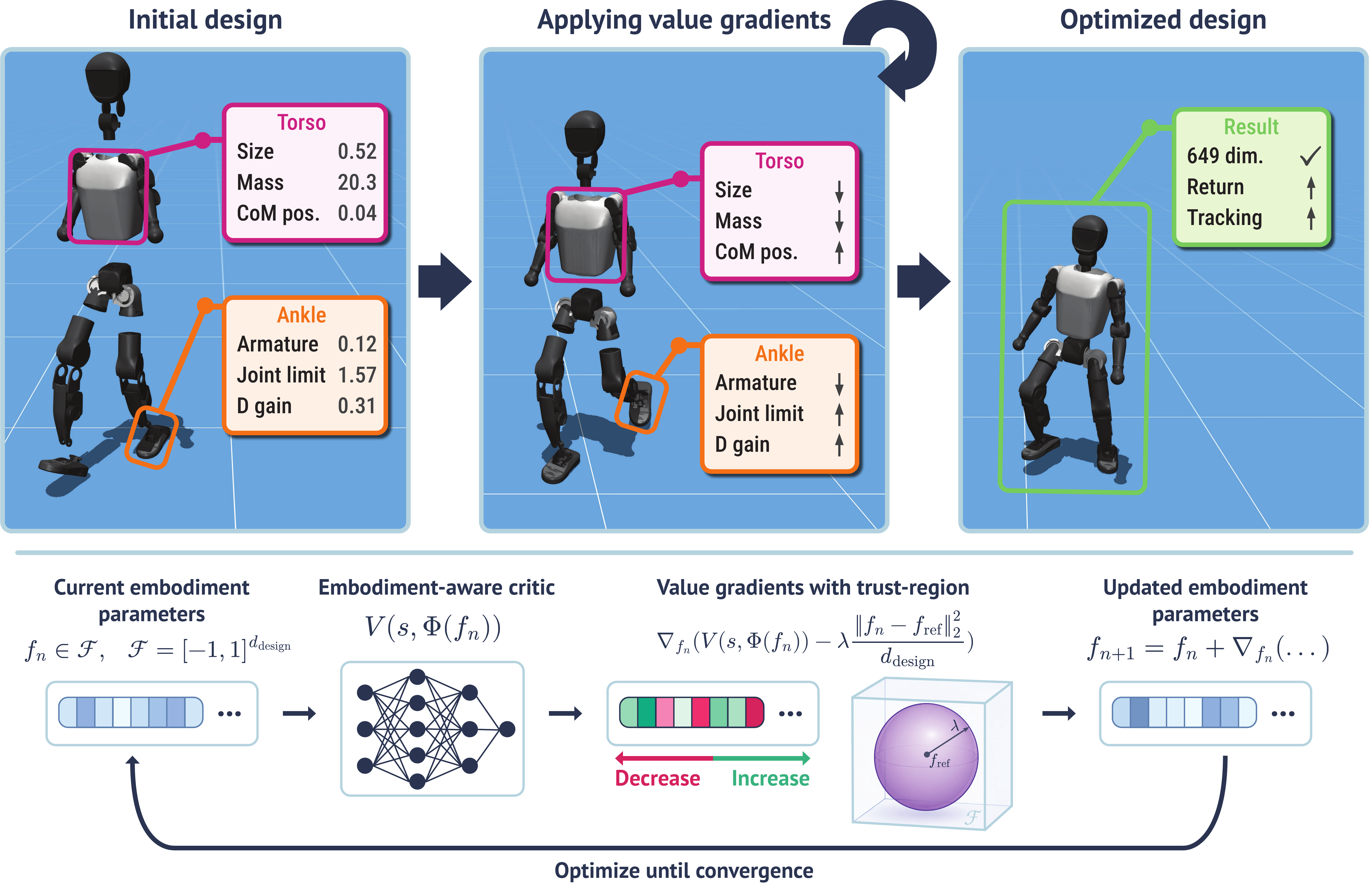}
\caption{
\textbf{Shape Your Body.}
We first train an embodiment-aware policy and value function with multi-embodiment reinforcement learning, then we optimize new designs by differentiating through the value function and applying the gradients inside a soft trust region around a reference design.
}
\label{fig:hero_figure}
\end{figure}

\begin{abstract}
We propose to turn generalist multi-embodiment value functions into reusable models for robot design.
Instead of running a new reinforcement learning co-design loop for each robot, we first train an embodiment-aware policy and value function across many robot designs.
After training, the frozen value function is used as a differentiable surrogate to optimize candidate embodiments through value gradients.
We evaluate our approach across different robot design settings, from perturbed single robots to held-out robots across morphology classes, with single models trained on up to 50 robots and design spaces of over 1100 continuous embodiment parameters.
Beyond optimizing complete embodiments, we show that value gradients can identify performance-limiting design and control parameters, enabling both the optimization and the analysis of new robot designs.
\end{abstract}



\section{Introduction}
Deep learning methods have revolutionized control and decision-making in robotics by enabling policies to learn complex behaviors in a data-driven manner.
In contrast, robot hardware is still largely designed by hand, relying on human engineering expertise.
With learning-based control and reliable simulation becoming important design considerations, this manual design process now depends on the intuition of both mechanical engineers and robot learning researchers.
As a robot's embodiment fully defines its physical capabilities, even an optimal policy can only be as good as the embodiment allows it to be.
Robot design and control are therefore naturally coupled, and jointly optimizing both is commonly referred to as robot co-design \cite{sims1994evolving, lipson2000automatic}.

Most co-design methods treat this coupling as a bi-level optimization problem.
An outer loop proposes embodiment designs, while an inner loop trains or adapts a controller for each candidate.
Evaluating each design often requires full \gls{rl} runs or well-tuned optimal control.
Recent \gls{rl}-based co-design methods reduce this cost by learning embodiment-aware policies and updating design and control within a single training process \cite{luck2020data,yuan2021transform2act}.
However, for every new robot, design space, or task setup, the complete co-design loop must be run again.
As a result, scaling co-design to many robots or many design variants remains difficult.
For real-world or realistically simulated robots, the design space itself is often high-dimensional, with actuator properties, body-part masses, inertias, and geometry easily reaching hundreds of continuous design parameters.

We instead build on recent progress in learning generalist robot policies across many embodiments \cite{bohlinger2024onepolicy, ai2025towards}.
Rather than treating every co-design problem as a new training run, we amortize the process by first training a single multi-embodiment \gls{rl} policy and value function over a wide distribution of robot designs.
The policy learns to control many embodiments, while the value function learns to evaluate how well the shared policy is expected to perform for a given state and embodiment.
Once trained, the policy and value function are frozen and reused for downstream design.
Because the value function has been trained across many embodiments and morphology classes, it can draw on information from the broader robot distribution when evaluating a candidate design, rather than relying only on data collected for the robot currently being optimized.
In addition, downstream design search is computationally cheap, as many candidate robots or design variations can be optimized in parallel through batched value function evaluations, without new expensive \gls{rl} training runs.

Our experiments study this idea on a robot locomotion task at increasing levels of generalization.
We first test our gradient-based design method in the simplest setting, where a value function trained on randomized versions of a single robot is used to recover better designs from perturbed variations.
In this setting, we compare against a broad set of derivative-free optimizers that all use the same frozen value function as their surrogate objective.
We then move to the more relevant setting where the target robot was not part of the training set, by training a policy and value function on many robots of a morphology class and using it to improve a held-out robot.
Finally, we train a single policy and value function across all 50 robots in our dataset, spanning quadrupeds, humanoids, and hexapods, and apply it to robot variations across the morphology classes.
This lets us study robot design in substantially higher dimensional spaces than typical robot co-design work, reaching over 1100 continuous embodiment parameters on realistic legged robot models.
Besides optimizing complete embodiments, we show that the same value function can be used for design analysis by using its gradients to identify performance-limiting embodiment parameters, such as actuator torque and velocity limits, or body-part masses.
These gradients can also provide insight into the control setup, for example by indicating how PD gains should be adjusted for a given robot.
Together, these results show that generalist multi-embodiment value functions can act as reusable design models, which we see as a first step toward interactive engineering tools for optimizing and understanding new robot designs.


\newpage
\section{Related Work}

\begin{figure}[t]
\centering
\includegraphics[width=\textwidth]{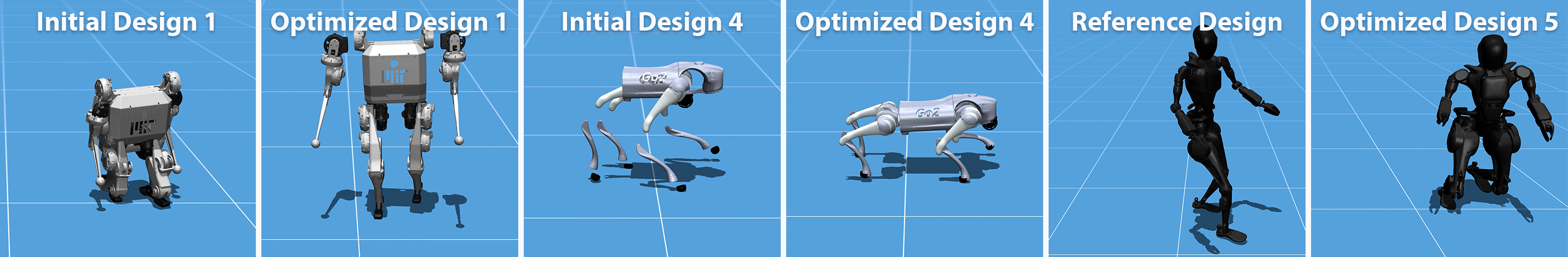}
\caption{
\textbf{Optimized robot designs.}
We visualize two initial designs and the corresponding optimized designs for the Go2 and the MIT Humanoid.
We also show the reference design and an optimized design for the Fourier GR1 T2 humanoid.
Although some assets appear stretched or visually disconnected, the underlying geometries remain connected and controllable by the policy, as discussed in \autoref{app:design_analysis_details}.
Full optimization trajectories across initial designs for our main three robots are shown in \autoref{app:design_trajectory_gallery}.
}
\label{fig:robots}
\vspace{-0.5em} 
\end{figure}

In biological systems, morphology and behavior co-develop, i.e., morphology determines which behaviors are possible, while behavioral pressures drive morphological change through evolution \cite{bertossa2011morphology}.
Robotics inherited this perspective, and through the lens of evolutionary algorithms, foundational work established the embodiment as an optimization variable instead of a fixed constraint when optimizing control \cite{sims1994evolving, lipson2000automatic, hornby2001body}.
Since then, many works have focused on discrete, graph-based, or grammar-based representations to search over robot structures \cite{wang2019neural, banarse2019body, zhao2020robogrammar, xu2021multi, hejna2021task, yuan2021transform2act, schaff2022n, qiu2024robomorph, gao2026vlmgineer}, while black-box optimization methods have been used to optimize continuous embodiment parameters or latent embodiment representations \cite{yu2022multi, hu2023glso, li2023learning, yi2025soft, ikemura2026latent, vaish2026identifying}.
These approaches established embodiment optimization as a core robotics problem, but evaluating each design typically requires either costly \gls{rl} training or potentially suboptimal classical control methods.

A more recent line of work formulates co-design directly as a \gls{rl} problem, enabling experience sharing across design candidates within a single training process \cite{ha2019reinforcement, schaff2019jointly, luck2020data, yuan2021transform2act, chen2021hardware, schaff2022n, wang2023curriculum, dong2023symmetry, li2024reinforcement, lu2025bodygen, dai2026stackelberg}.
Closest to our use of a value function for design search, \gls{feacrl} \cite{luck2020data} uses the state-action value function learned by \gls{sac} \cite{haarnoja2018soft} to continuously score candidate designs, which are selected with \gls{pso} \cite{kennedy1995pso}.
Starting with Transform2Act \cite{yuan2021transform2act}, most recent \gls{rl}-based methods split episodes into a design phase and a control phase, with dedicated subpolicies.
BodyGen \cite{lu2025bodygen} improves the phase-specific credit assignment and adds topology-aware representations, while Stackelberg \gls{ppo} \cite{dai2026stackelberg, schulman2017proximal} models the leader-follower dynamics of the two subpolicies.
These methods improve the efficiency of the co-design process, but their optimization remains tied to the individual robot and its co-design run.
In contrast, we first train the policy and value function over a broad robot distribution, allowing downstream search to use cross-embodiment information while reusing the value function to design many robots or design variants without retraining.

In embodiment-aware learning, the policy and value function can reason about the robot's physical embodiment rather than treating it as a hidden variable.
Multi-embodiment learning is a common approach, where a single policy and value function are explicitly or implicitly conditioned on embodiment information.
Early methods used \glspl{gnn} to model the robot's kinematic tree \cite{wang2018, huang2020one}, while more recent transformer-based architectures treat joints, actuators, or body parts as tokens to improve scalability \cite{yu2022multi, gupta2022, patel2025get, sferrazza2025body}.
Other approaches infer the embodiment from proprioceptive history \cite{liu2025locoformer, li2026online}, and on-robot learning methods directly train on the real embodiment \cite{smith2023walk, smith2024grow, levy2024learning, bohlinger2025gait}.
Recent large-scale multi-embodiment approaches have shown that generalist policies can be trained across broad robot distributions with architectures such as the \gls{urma} \cite{bohlinger2024onepolicy, ai2025towards, bohlinger2025multi, li2026online}.
We build on this by using \gls{urma} to train a generalist embodiment-aware policy and value function.
While prior multi-embodiment work focuses on using the trained policy for control or for evaluating new designs \cite{strgar2026accelerated, fay2026dextra}, we use the embodiment-aware value function after training as a reusable surrogate for design optimization and analysis.

Our design search is related to commonly used optimization methods for robot co-design.
Classical black-box optimizers such as \gls{bo} \cite{balandat2020botorch}, \gls{cmaes} \cite{hansen2001completely}, \gls{pso} \cite{kennedy1995pso}, \gls{cem} \cite{rubinstein2004cross}, \gls{de} \cite{storn1997differential}, \gls{ars} \cite{mania2018simple}, and \gls{turbo} \cite{eriksson2019scalable} are commonly used when gradients of the objective are unavailable or unreliable.
Recent gradient-based population methods such as \gls{gcpfo} \cite{vaish2026identifying} use the gradient covariance to guide the search in high-dimensional robot design spaces.
Our design search operates directly on the trained embodiment-aware value function and therefore has access to the analytic gradients with respect to the high-dimensional embodiment parameters.
This is also closely related to explicit policy-conditioned value functions, which learn universal critics conditioned on policy representations, like parameters or embeddings, and update policies by differentiating through the critic with respect to these representations \cite{harb2020, faccio2021, faccio2022, bohlinger2025massively}.


\section{Value Gradients for Multi-Embodiment Robot Design}
\label{sec:method}

We propose to turn the embodiment-aware value function, obtained from large-scale multi-embodiment \gls{rl} training, into a reusable surrogate for robot design.
The core idea is shown in \autoref{fig:hero_figure}.
Instead of running a new co-design \gls{rl} loop for every target robot, we first train a single embodiment-aware policy and critic across many robot designs.
After training, the policy and critic are frozen.
The critic is then differentiated with respect to robot design parameters and reused as a design model, so many target embodiments can be optimized with the same frozen networks.

We study robot design for robots with a given kinematic structure, where continuous physical parameters, such as masses, inertias, geometry, joint limits, and actuator properties, can be modified.
Each robot is described by a continuous embodiment vector $e \in \mathbb{R}^{d_{\mathrm{phys}}}$ that determines the transition dynamics $p_e(s_{t+1}\mid s_t,a_t)$ and reward function $r_e(s_t,a_t)$.
For an embodiment-aware policy $\pi_\theta(a_t\mid s_t,e)$ with parameters $\theta$, the expected discounted return on embodiment $e$ is
\begin{equation}
J(\theta,e)
=
\mathbb{E}_{\tau\sim\pi_\theta(\cdot\mid\cdot,e)}
\left[
\sum_{t=0}^{T}\gamma^t r_e(s_t,a_t)
\right],
\label{eq:return}
\end{equation}
where $s_t$ and $a_t$ are the state and action at time $t$, $\tau=(s_0,a_0,\ldots,s_T)$ is the trajectory generated by rolling out $\pi_\theta$ under the dynamics $p_e$, $T$ is the episode horizon, and $\gamma$ is the discount factor.

We represent each candidate robot design in a normalized design space $\mathcal{F}=[-1,1]^{d_{\mathrm{design}}}$ and let
$f\in\mathcal{F}$ be the normalized design vector.
The reference design $f_{\mathrm{ref}}\in\mathcal{F}$ serves as the anchor for the trust region and can be any valid design, such as a nominal URDF for an existing robot or a completely new one.
A differentiable map
$e = \Phi(f)$
converts the design from the normalized space to physical embodiment parameters such as joint origins and ranges, actuator force and velocity limits, damping, stiffness, armature, friction, PD gains, body masses, inertias, center-of-mass offsets, link geometry, or foot parameters.
Hard physical bounds inside $\Phi$ can be used to rule out physically implausible or unmanufacturable designs, such as negative masses or overlapping body parts.
All design parameters, their ranges, and the constraints that we use are listed in \autoref{app:design_space_table}.
Design-specific reward terms can also be added to the reward function to discourage plausible but undesirable design features \cite{ha2019reinforcement}; however, we omit this in our setting.
The goal of the design process is to improve an initial design $f_{\mathrm{init}} \in \mathcal{F}$ by iterating on it and finding a design $f^\star \in \mathcal{F}$ with higher return $J(\theta,\Phi(f^\star))$ when rolled out with the frozen policy $\pi_\theta$.

\subsection{Training a Multi-Embodiment Policy and Critic}
\label{sec:method_training}

During \gls{rl} training, each parallel environment $i$ samples a new design
$f_i \sim \mathcal{U}([-c_t,c_t]^{d_{\mathrm{design}}})$
at an episode reset, maps it to a physical robot $e_i = \Phi(f_i)$, and keeps that design fixed for the episode.
Designs can come from multiple base robots of different morphology classes.
The performance-based curriculum coefficient $c_t$ expands the design support during training and reaches the full design space at $c_t=1$ \cite{bohlinger2025multi}.
The policy is trained with \gls{ppo} to maximize
\begin{equation}
\max_\theta
\;
\mathbb{E}_{f\sim\mathcal{U}([-1,1]^{d_{\mathrm{design}}})}
\left[
J(\theta,\Phi(f))
\right].
\end{equation}

We use \gls{urma} as our policy and critic architecture to handle the varying topologies, kinematics, and dynamics across our multi-embodiment training set.
For a robot with a set of joints $\mathcal{J}$, \gls{urma} splits observations into fixed-size general observations $o_g$ and variable-size per-joint observations $\{o_j\}_{j\in\mathcal{J}}$.
Each joint is paired with a description vector $d_j$ that encodes static joint and adjacent body properties, such as joint axis, relative position, limits, gains, body mass, inertia, and geometry.
The same idea is used for feet observations, which are present only in the critic network.
\gls{urma} encodes each joint separately and aggregates the variable number of latents with an attention scheme:
\begin{equation}
\bar z_{\mathrm{joints}}
=
\sum_{j\in\mathcal{J}}
\alpha_j(d_j) \odot g_\psi(o_j),
\qquad
\alpha_j(d_j)
=
\frac{
\exp\!\left(g_\phi(d_j)/\eta\right)
}{
\sum_{L_d}
\exp\!\left(g_{\phi}(d_j)/\eta\right)
},
\label{eq:urma_attention}
\end{equation}
where $g_\psi$ and $g_\phi$ are learned encoders, $\eta$ is a learnable temperature, $L_d$ is the latent dimension, and $\odot$ denotes element-wise multiplication.
The aggregated joint latent $\bar z_{\text{joints}}$, foot latent $\bar z_{\text{feet}}$, and general observations $o_g$ are then processed by a shared core network.
For the policy, the resulting core latent is decoded back into one action distribution per joint.
For the critic, the core latent is decoded into the scalar value prediction.

We modify the standard \gls{urma} critic into a direct-design \gls{urma} critic to make its value predictions more sensitive to the design parameters.
In the original critic, the description vectors $d_j$ only serve as input to the attention keys, while the attention values are computed from the per-joint observations $o_j$.
We extend the encoder $g_\psi$ to take both the observations and the descriptions as input, $g_\psi(o_j, d_j)$, so the embodiment parameters have a more direct influence on the attention values and therefore on the final discounted return prediction, which leads to empirically stronger design gradients.
To reduce the sensitivity to approximation errors in the value function, which could lead to catastrophic design updates, we use an ensemble of $K$ core-value heads on top of shared per-joint and per-foot encoders, and optimize their mean prediction.
An architecture overview and the ablations that led to the direct-design critic are summarized in \autoref{app:critic_architecture}.

\subsection{Value-Gradient Design Search}
\label{sec:method_search}

After multi-embodiment training, we freeze the policy and critic and run \acrfull{vgds} to optimize the normalized design vector $f$ using gradients of the learned value function.
Let $V_k(s,\Phi(f))$ denote the value prediction of critic head $k$ for state $s$ and design $f$.
We combine the $K$ heads into the mean value prediction
\begin{equation}
\bar V(s,\Phi(f)) = \frac{1}{K}\sum_{k=1}^{K} V_k(s,\Phi(f)).
\end{equation}

To make the design search cheap and deterministic, we first roll out the final policy on the full design space $\mathcal{F}$ to collect a diverse state bank ${\mathcal{S}=\{s_1,\dots,s_M\}}$.
The critic is trained on designs visited during the multi-embodiment training phase, so its predictions can be unreliable when the search moves too far from the experienced distribution.
In preliminary experiments, unconstrained critic maximization often moved designs toward sub-spaces of the design space, where the critic predicted high returns but the real performance collapsed, which we analyze in \autoref{app:critic_reliability}.
We therefore restrict the design search to a trust region around a provided reference design $f_{\mathrm{ref}}$:
\begin{equation}
\max_{f\in\mathcal{F}}
\;
\mathbb{E}_{s\sim\mathcal{U}(\mathcal{S})}
\left[
\bar V(s,\Phi(f))
\right]
\quad
\text{s.t.}
\quad
\frac{\left\|f-f_{\mathrm{ref}}\right\|_2^2}{d_{\mathrm{design}}}
\leq
\epsilon.
\label{eq:trust_region_problem}
\end{equation}
We relax the constraint into a penalty with a fixed coefficient $\lambda$, which gives our main design optimization objective:
\begin{equation}
\hat J_\lambda(f)
=
\mathbb{E}_{s\sim\mathcal{U}(\mathcal{S})}
\left[
\bar V(s,\Phi(f))
\right]
-
\lambda
\frac{\left\|f-f_{\mathrm{ref}}\right\|_2^2}{d_{\mathrm{design}}}.
\label{eq:soft_trust_region}
\end{equation}

The quadratic penalty discourages designs that move far away from the reference robot, where the critic is more likely to extrapolate.
The normalization by $d_{\mathrm{design}}$ ensures that the same $\lambda$ has a comparable effect across robots with differently sized design spaces.
The optimizer can still move far from the reference when the gain in predicted value is large enough to compensate for the penalty.
We use $\hat J_\lambda(f)$ as the objective for \gls{vgds} and for all other search baselines.

\gls{vgds} optimizes \autoref{eq:soft_trust_region} by gradient ascent and estimates the expectation with a minibatch ${B\subset\mathcal{S}}$ of the state bank in every iteration.
This leads to the design gradient
$
g_n
=
\nabla_f \hat J_{\lambda}(f_n)
$
at iteration $n$, which contains a value gradient term obtained by backpropagation through the frozen critic and the design map $\Phi$, and a second term through the analytic trust-region gradient ${-2\lambda (f_n-f_{\mathrm{ref}}) / d_{\mathrm{design}}}$.
We then apply an Adam \cite{kingma2015adam} ascent step, clip the update of each design parameter to at most $\delta_{\max}$, and finally clip the updated design to the valid design space $\mathcal{F}$:
\begin{equation}
f_{n+1}
=
\mathrm{clip}_{[-1,1]}
\left(
f_n
+
\mathrm{clip}_{[-\delta_{\max},\delta_{\max}]}
\left[
\Delta_{\mathrm{Adam}}(g_n)
\right]
\right).
\label{eq:soft_update}
\end{equation}

\gls{vgds} finds the final design $f^\star = f_N$ after $N$ iterations of applying \autoref{eq:soft_update}.
Finally, additional variants of \gls{vgds} that we tested during the development are summarized in \autoref{app:additional_method_variants}.


\section{Experiments}
\label{sec:experiments}

To evaluate our method on a high-dimensional design setting with realistic robot models, we use a standard sim-to-real-transferable velocity-tracking locomotion task, where each robot has to follow randomly sampled $x$-$y$ linear and yaw angular velocity commands, from previous \gls{urma}-based multi-embodiment learning work \cite{bohlinger2024onepolicy, ai2025towards, bohlinger2025multi}, implemented in RL-X \cite{bohlinger2023rlx} with \gls{mjx} \cite{todorov2012mujoco}.
Across experiments, the policy and critic are trained on distributions drawn from up to 50 base robots spanning 15 quadrupeds, 31 bipeds and humanoids, and 4 hexapods.
Further details on the environment, the set of robots, and the RL training setup are given in \autoref{app:experimental_setup}.

We report and compare the different methods using the return improvement
$\Delta R = R(f^\star) - R(f_{\mathrm{init}})$,
where $f_{\mathrm{init}}$ is the design at which search starts and $f^\star$ is the final design found by the optimizer.
To also compare the results on a more robotics-relevant metric of the task, we report the reduction in tracking error of the $x$-$y$ linear velocity command in \autoref{app:velocity_tracking_results}.

\subsection{Single-Robot Design}
\label{sec:single_robot_design}

\begin{figure}[t]
\centering
\includegraphics[width=\textwidth]{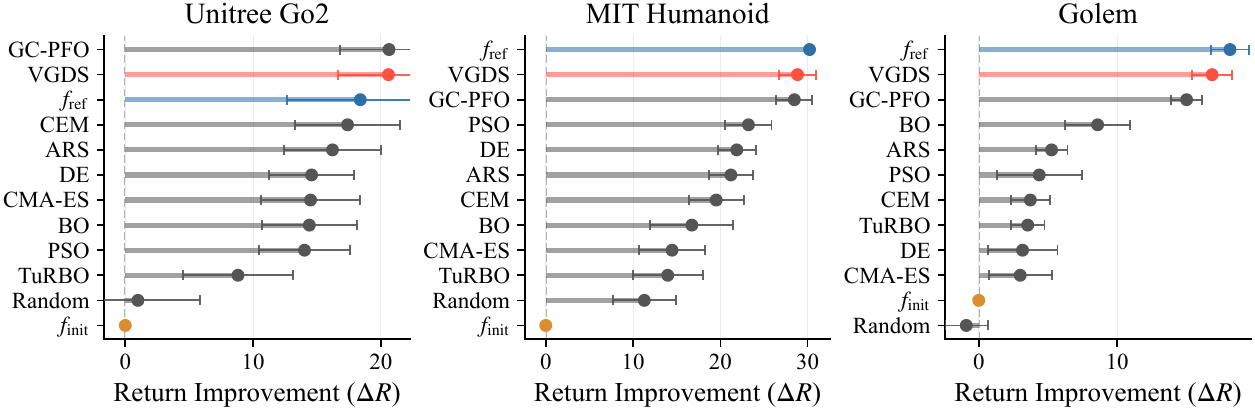}
\caption{
\textbf{Single-robot design.}
Mean return improvement $\Delta R$ over the initial perturbed design $f_{\mathrm{init}}$ for $10$ starts per robot.
The nominal URDF design $f_{\mathrm{ref}}$ is shown as a reference, but is only used as the anchor for the trust region.
}
\label{fig:e1_return}
\vspace{-0.5em} 
\end{figure}

\begin{figure}[t]
\centering
\includegraphics[width=\textwidth]{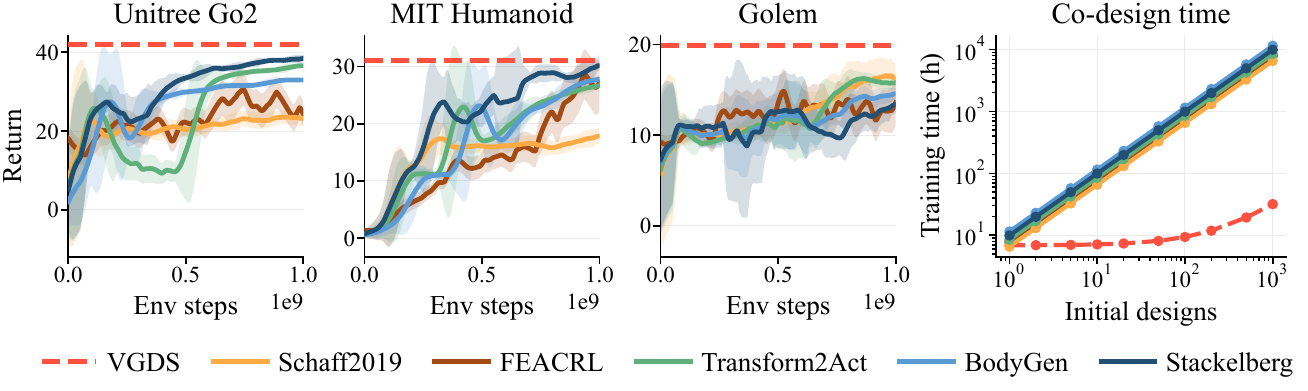}
\caption{
\textbf{Comparison to RL-based co-design.}
The first three plots show the return over environment steps for Schaff2019 \cite{schaff2019jointly}, \gls{feacrl}, Transform2Act, BodyGen, and Stackelberg PPO.
The dashed line shows the final performance of \gls{vgds} after training a policy and critic on the full design space, and then designing from the same $f_{\mathrm{init}}$ as the \gls{rl} baselines started from.
The fourth plot shows the cumulative time needed to co-design increasing numbers of initial designs.
}
\label{fig:rl_baselines}
\vspace{-1.0em} 
\end{figure}

We first evaluate our design algorithm without cross-robot generalization.
We train three separate embodiment-aware policies and critics for uniformly sampled random designs from the full design space $\mathcal{F}$ for the Unitree Go2 quadruped with $358$ embodiment parameters, the MIT Humanoid with $514$ embodiment parameters, and the Golem hexapod with $688$ embodiment parameters.
We then start to design $10$ uniformly sampled designs from $\mathcal{F}$ for each robot and optimize their embodiment parameters.
As a first investigation, we ablate the effect of the trust-region penalty coefficient $\lambda$ in \autoref{app:exp_trust_region_ablation}, and choose $\lambda=100$ for all subsequent experiments.
Supplementary results on the effect of the trust region and other design choices are also provided in \autoref{app:exp_trust_region_ablation}.

\autoref{fig:e1_return} shows that \gls{vgds}, like the other search methods, can significantly improve the sampled designs $f_{\mathrm{init}}$ across all three robots, and comes close to, or even exceeds, the performance of the robot's nominal design $f_{\mathrm{ref}}$.
$f_{\mathrm{ref}}$ most often acts as a strong performance reference, as it is an already well-engineered design and all randomizations are sampled from the design space centered around it.
On the Unitree Go2 and MIT Humanoid, \gls{vgds} matches the strongest baseline, and on Golem, it gives the largest improvement of all methods.
Across all experiments, \gls{gcpfo} proves to be the strongest baseline, as it is the only other method that leverages gradient information, which shows the effectiveness of using the value gradients in such high-dimensional design spaces.
Additionally, fine-tuning the policy on the found designs can further improve its performance, which we analyze in \autoref{app:policy_finetune}.

We also compare \gls{vgds} to \gls{rl}-based co-design methods that train a new policy and find a new design for each $f_{\mathrm{init}}$.
We substantially adapt the baselines to our setting using \gls{urma}-style policies and critics to handle the different morphologies and high-dimensional embodiment parameters, and train them all with \gls{ppo}.
Further details about the adaptations of the baselines to our setting are given in \autoref{app:rl_baseline_adaptations}.
\autoref{fig:rl_baselines} shows that \gls{vgds} reaches performance on par with or slightly better than the adapted \gls{rl} baselines.
This difference in training efficiency becomes important when many initial designs have to be optimized.
The adapted \gls{rl} baselines scale linearly with the number of designs, as every initial design requires a separate training run, whereas our approach reuses the same policy and critic.
After the initial training run of about $7$--$9$ hours, designing each additional design takes only about $1$--$2$ minutes of search time.
Most of this time comes from initialization processes like JIT compilation and can be further reduced by simply batching the critic inference of multiple designs together.

\subsection{Generalization Across Training Distributions}
\label{sec:exp_generalization}

Next, we evaluate cross-embodiment generalization of \gls{vgds} by training the policy and critic on multiple base robots with the usual embodiment randomization on top.
We create training sets for each morphology class but remove the target robot from the training set.
Finally, we create one training set that includes all 50 robots, including the target robot.
Both settings use the same uniformly sampled random initial designs $f_{\mathrm{init}}$ for the design process, which are sampled from the full design space $\mathcal{F}$ of the target robot, as in the previous experiment.

\begin{figure}[t]
\centering
\includegraphics[width=\textwidth]{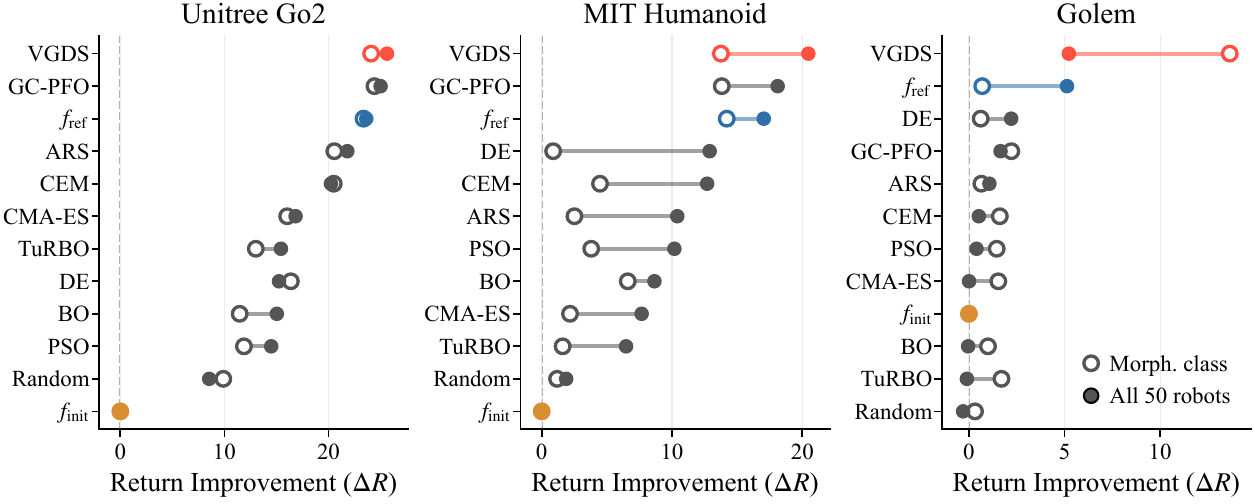}
\caption{
\textbf{Effect of training sets.}
Target robot is either held out from a morphology class set (open circles) or included in the full 50-robot training set (filled circles).
We omit error bars for readability and report them in \autoref{app:numeric_results}.
}
\label{fig:e2_e4_return}
\vspace{-0.8em}
\end{figure}

\autoref{fig:e2_e4_return} shows that \gls{vgds} remains the strongest search method.
When trained on all 50 robots, \gls{vgds} can find designs that reach higher performance than $f_{\mathrm{ref}}$ for the Go2 and the Humanoid, which is a significant improvement over the single-robot training and shows the benefit of training on a broader distribution of robots.
For Golem, the trend is different, as the search methods using the hexapod-only training set reach higher improvement than the ones trained on all robots.
This is likely due to the hexapod class consisting of only 4 robots, which means the full training set is dominated by quadrupeds and bipeds, and the critic trained on all robots is less specialized for the hexapods.
To verify this, we train an additional class-balanced model on 12 robots and rerun the design experiments in \autoref{app:balanced_training}.
Additionally, we provide the design results for all 50 robots in \autoref{app:full_robot_design_results}.

\subsection{Design Analysis with Value Gradients}
\label{sec:exp_design_analysis}

\begin{figure}[t]
\centering
\includegraphics[width=\textwidth]{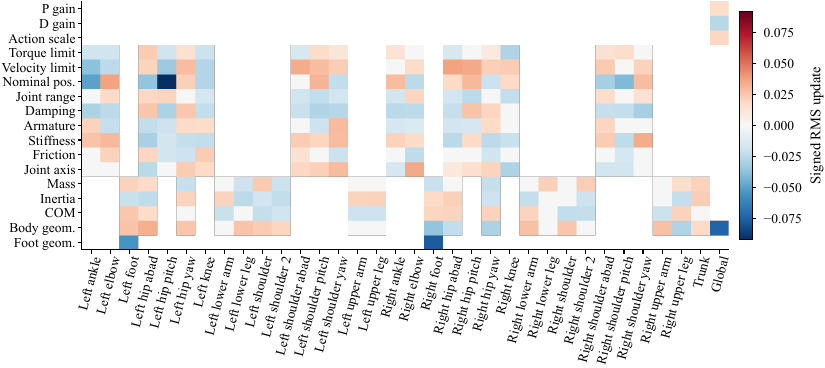}
\caption{
\textbf{Grouped design changes for the MIT Humanoid.}
Signed \acrfull{rms} changes of the optimized designs by \gls{vgds} relative to the nominal reference, grouped by body part and parameter type.
This can give an overview over which parameters are most influential for the design optimization.
The heatmaps for the Unitree Go2 and Golem are shown in \autoref{app:design_analysis_details}.
}
\label{fig:design_analysis_heatmap_mit}
\end{figure}

We inspect what \gls{vgds} changes by grouping the updates $f^\star-f_{\mathrm{ref}}$ by body part and parameter type.
Besides physical parameters such as masses or geometry, we look at how \gls{vgds} can also provide insights into tuning control parameters.
For the MIT Humanoid, the overall strongest changes are nominal joint positions and gains, together with reduced foot size (see \autoref{fig:design_analysis_heatmap_mit}).
Copying only the optimized gains into the initial design improves the return already from $5.8$ to $12.5$ (see \autoref{fig:design_analysis_counterfactuals}).
For Golem, \gls{vgds} significantly reduces the action scale, lowers the P gain, and increases the D gain.
For the Unitree Go2, the most prominent changes are more physical and local, including rear leg joint axis changes, foot geometry changes, and different actuator velocity limits on the front hip and calf.
Additional evaluations in \autoref{app:design_analysis_details} show that no single parameter group explains the full improvement, and the final designs come from coupled updates across many parameters.
Lastly, \autoref{fig:robots} shows an interesting observation for many tall humanoids, such as Fourier GR1 T2, where \gls{vgds} makes the robots visibly smaller, wider, and more compact than the reference design, which appears to improve stability and may move them closer to the center of the humanoid training distribution.

\section{Limitations}

\gls{vgds} currently optimizes continuous parameters of a fixed robot topology and does not add or remove joints or body parts, as there is no gradient for topology changes through the \gls{urma} network.
The method also depends on the coverage and accuracy of the policy and value function.
While the trust region helps with errors in extrapolation, having a sufficiently good reference design available is not always possible.
Integrating fine-grained design priors or doing the design search in a latent space might be ways to mitigate this.
Finally, all experiments are performed in rather simple MuJoCo simulations.
We therefore do not evaluate whether the optimized designs are manufacturable or transferable to hardware.
The training pipeline itself is sim-to-real transferable, as shown in \autoref{app:sim2real}.
Integrating manufacturability properly would require additional constraints or simulators for materials, electronics, actuation, and fabrication, which continues to be an open challenge for robot co-design.


\section{Conclusion}

We introduced \emph{Shape Your Body}, a method for amortized robot design leveraging gradients from generalist multi-embodiment value functions.
After training a policy and value function across many randomized robot embodiments, \gls{vgds} reuses the value function as a surrogate design model and optimizes hundreds of continuous embodiment parameters within minutes.
Across quadrupeds, humanoids, and hexapods, \gls{vgds} significantly improves sampled designs, and matches or exceeds adapted \gls{rl} co-design baselines while having a much lower marginal cost per additional design.
Beyond performance optimization, we see a path toward interactive co-design tools that can optimize a candidate robot and point engineers to relevant changes in gains, actuator limits, geometry, and other parameters.
Future work should extend to topology changes, the integration of more complex design constraints in $\Phi(f)$, task-specific design objectives, and trust regions in latent spaces.


\clearpage
\acknowledgments{
This project was funded by National Science Centre Poland in the Weave programme UMO-2021/43/I/ST6/02711, and by the German Science Foundation (DFG) under grant number PE 2315/17-1.
}


\bibliography{references}

\clearpage

\appendix

\section*{Appendix}

\section{Experimental Setup Details}
\label{app:experimental_setup}

\subsection{Environment}
\label{app:setup_env}

All robots are simulated in a fully JAX-jitted \gls{mjx} locomotion environment implemented in RL-X.
The environment runs at a $200$~Hz simulation frequency with a $50$~Hz control frequency, and each episode lasts at most $20$~s, corresponding to $1{,}000$ control steps.
Episodes terminate early if the root height drops below $80\%$ of the nominal stance height, which is linearly scaled to $0\%$ for a curriculum coefficient of $1.0$.
We use the same environment setup for all robots, where the only difference is the nominal joint positions and PD gains, while everything else is shared and automatically adjusted to each robot, e.g. through size-, mass-, or number-of-joints-based scaling of rewards or randomization ranges.

\paragraph{Actions and PD controller.}
The policy outputs clipped actions $a_j \in [-10,10]$ for each actuated joint.
Actions are converted to target joint positions as
\begin{equation}
q^{\mathrm{target}}_j = q^{\mathrm{nom}}_j + \sigma_{\mathrm{robot}} a_j,
\end{equation}
where $q^{\mathrm{nom}}_j$ is a nominal joint position and $\sigma_{\mathrm{robot}}$ is the robot-specific action scaling factor listed in \autoref{tab:robots_all}.
The target positions are tracked by a standard PD controller using the p- and d-gains listed in the same table.
The gains and action scaling factor are also part of the searchable design space.

\begin{figure}[t]
\centering
\includegraphics[width=\textwidth]{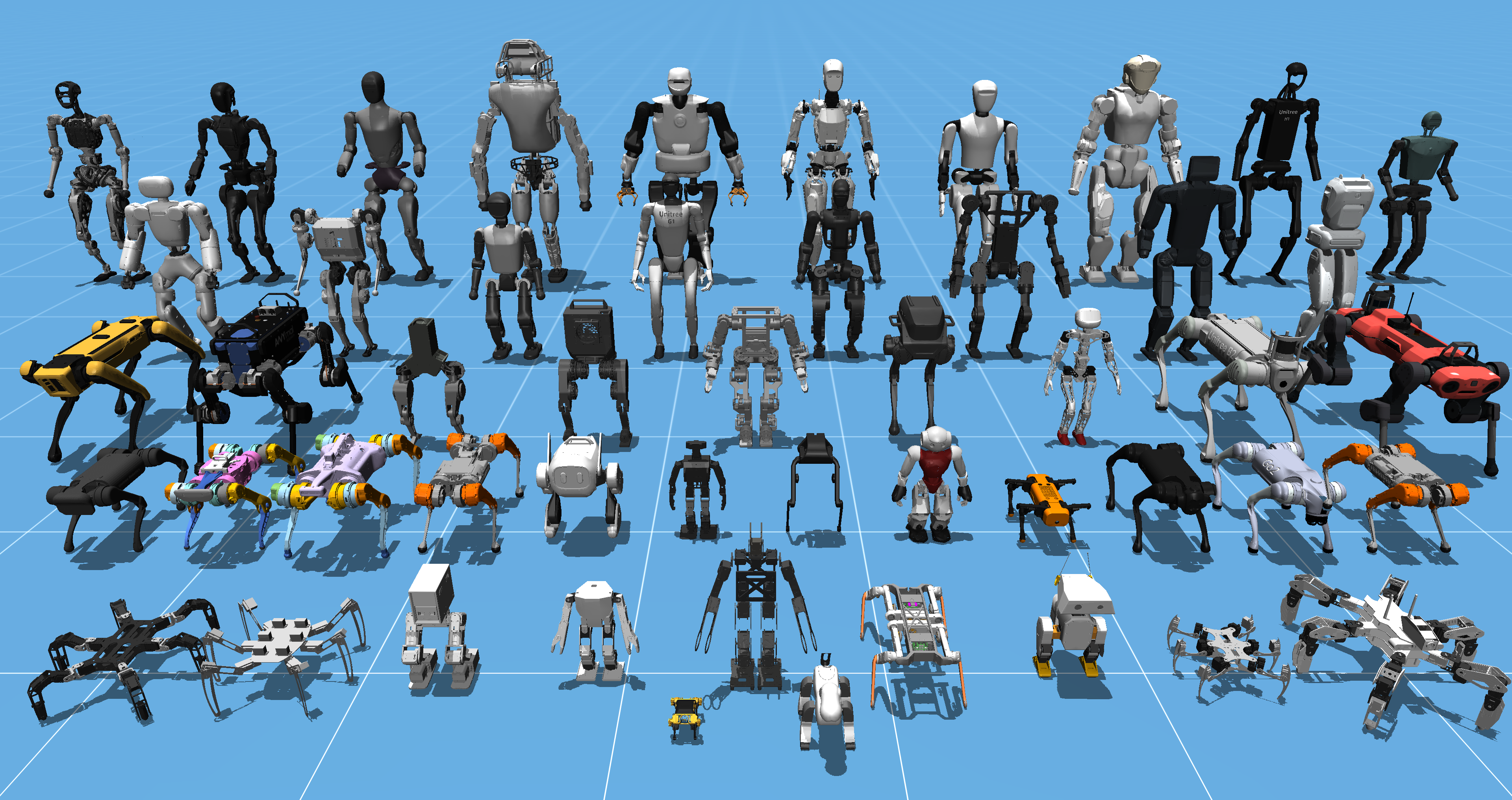}
\caption{
\textbf{Overview of all 50 robots used in the multi-embodiment RL training \cite{bohlinger2025multi}.}
}
\label{fig:robot_overview}
\end{figure}

\paragraph{Observations and descriptions.}
The observations of each actuated joint $o_j$ consist of its position relative to the nominal joint position, velocity, previous action, and an indicator for whether the joint should keep the nominal position, e.g., arms of humanoids or all joints when commanded velocities are zero.
The observations of each foot $o_f$ consist of a contact flag, time on ground, and time in air, and are only used by the critic.
Static embodiment information is encoded through per-joint and per-foot description vectors $d_j$ and $d_f$.
Each joint description $d_j$ contains the relative joint position, joint axis, attached-body mass, attached-body center of mass, attached-body inertia, number of direct child actuator joints, nominal joint position, torque and velocity limits, damping, armature, stiffness, friction loss, joint range, PD gains, action scaling factor, total robot mass, and robot dimensions.
Each foot description $d_f$ contains the relative foot position, foot-body mass, foot-body center of mass, foot-body inertia, foot geometry size and type, PD gains, action scaling factor, total robot mass, and robot dimensions.
The policy additionally receives general observations such as the IMU angular velocity, command velocities, projected gravity, PD gains, action scaling factor, total mass, robot dimensions, and trunk mass, trunk center-of-mass position, and trunk inertia.
The critic also observes the IMU linear velocity, terrain height, and the current curriculum coefficient.

\subsection{Reward function}
\label{app:setup_reward}

The reward consists of $x$-$y$-$\mathrm{yaw}$ velocity-tracking terms ($T_1$ and $T_2$), alive rewards ($T_3$ and $T_4$), and regularization penalties ($T_5$ to $T_{23}$).
Let $c_t \in [0,1]$ be the curriculum coefficient, and let $T_i$ and $w_i$ denote the reward terms and coefficients in \autoref{tab:reward_terms}.
The unclipped pre-reward is
\begin{equation}
\tilde r_t =
w_1 T_1 + w_2 T_2
+ c_t \left(
w_3 T_3 + \sum_{i=5}^{23} w_i T_i
\right),
\end{equation}
and the final reward is
\begin{equation}
r_t = \max(\tilde r_t, 0) + c_t w_4 T_4.
\end{equation}
All coefficients are multiplied by the control timestep $\Delta t$.

\begin{table}[h]
\centering
\def\arraystretch{1.25}
\caption{
\textbf{Reward terms for locomotion training.}
Joint-based and foot-based terms are averaged over the corresponding actuated joints or feet to keep the reward scales consistent across different morphologies.
}
\label{tab:reward_terms}
\small
\begin{tabular}{l l l r}
\toprule
& Term & Equation & Coef. \\
\midrule
T1  & $x$-$y$ velocity tracking
    & $\exp(-\|v_{xy}-c_{xy}\|_2^2 / 0.25)$
    & $2.0$ \\
T2  & $\mathrm{yaw}$ velocity tracking
    & $\exp(-(\omega_z-c_{\omega})^2 / 0.25)$
    & $1.0$ \\
T3  & Alive clipped
    & $1$
    & $0.05$ \\
T4  & Alive unclipped
    & $1$
    & $0.05$ \\
T5  & Z velocity penalty
    & $-|v_z|^2$
    & $2.0$ \\
T6  & IMU acceleration penalty
    & $-\|\Delta v_{\mathrm{imu}} / \Delta t\|_2^2$
    & $10^{-4}$ \\
T7  & Roll-pitch velocity penalty
    & $-\|\omega_{xy}\|_2^2$
    & $0.05$ \\
T8  & Roll-pitch position penalty
    & $-\|\theta_{xy}\|_2^2$
    & $10.0$ \\
T9  & Joint nominal deviation penalty
    & $-\overline{\|q_u-q^{\mathrm{nom}}_u\|^2}$
    & $20.0$ \\
T10 & Joint position limit penalty
    & $-\overline{[\ell_u-q_u]_+ + [q_u-u_u]_+}$
    & $40.0$ \\
T11 & Joint velocity limit penalty
    & $-\overline{[|\dot q_u|-0.9\dot q^{\max}_u]_+}$
    & $5.0$ \\
T12 & Joint velocity penalty
    & $-\overline{\|\dot q_u\|^2}$
    & $4\times 10^{-4}$ \\
T13 & Joint acceleration penalty
    & $-\overline{\|\Delta \dot q_u / \Delta t\|^2}$
    & $5\times 10^{-6}$ \\
T14 & Joint torque penalty
    & $-\overline{\|\tau_u\|^2}$
    & $4\times 10^{-4}$ \\
T15 & Positive power penalty
    & $-\overline{[\tau_u \dot q_u]_+}$
    & $4\times 10^{-4}$ \\
T16 & Action rate penalty
    & $-\overline{\|a_t-a_{t-1}\|^2}$
    & $3.0$ \\
T17 & Action smoothness penalty
    & $-\overline{\|a_t-2a_{t-1}+a_{t-2}\|^2}$
    & $0.1$ \\
T18 & Self-collision penalty
    & $-n_{\mathrm{col}}$
    & $2.0$ \\
T19 & Base-height penalty
    & $-|h-h^{\mathrm{nom}}|^2$
    & $30.0$ \\
T20 & Foot air-time penalty
    & $\overline{\mathbbm{1}_{c_f}\min(t_f^{\mathrm{air}}-0.4s_{\mathrm{robot}},0)}$
    & $3.0$ \\
T21 & Symmetric air penalty
    & $-\overline{\mathbbm{1}_{\neg c_{f_l}}\mathbbm{1}_{\neg c_{f_r}}}$
    & $1.0$ \\
T22 & Foot slip penalty
    & $-\overline{\mathbbm{1}_{c_f}\|v^{f}_{xy}\|^2}$
    & $0.1$ \\
T23 & Foot z-velocity penalty
    & $-\overline{\|\min(v^{f}_z,0)\|^2}$
    & $0.2$ \\
\bottomrule
\end{tabular}
\end{table}

\subsection{Curriculum and design sampling}
\label{app:setup_curriculum}

At each episode reset, every training robot samples a normalized design vector
$f \sim \mathcal{U}([-c_t,c_t]^{d_{\mathrm{design}}})$ and keeps this design fixed for the episode.
Training starts with a curriculum coefficient of $c_t=0$, where only the nominal design is sampled, and expands toward $c_t=1$, where the full design space is available.
The curriculum follows the performance-based embodiment randomization scheme of Bohlinger et al.~\cite{bohlinger2025multi}.
All reported evaluation results always use the full design space.

We additionally randomize initial states and apply observation and actuation perturbations during training.
This includes a random action delay of up to one control step with $5\%$ mixed-delay probability, joint and IMU observation noise, random initial yaw, small roll-pitch perturbations, joint position perturbations around the nominal pose, and trunk velocity pushes.

\subsection{Robots and design dimensions}
\label{app:setup_robots}

\autoref{fig:robot_overview} shows the $50$ robots used in the largest training set, which contains $15$ quadrupeds, $31$ bipeds and humanoids, and $4$ hexapods.
For each robot, \autoref{tab:robots_all} reports the default actuator p-gain and d-gain, action scaling factor $\sigma_{\mathrm{robot}}$, number of actuated joints $n_u$, and design dimension $d_{\mathrm{design}}$, which is derived in \autoref{app:design_space_table}.


\begin{longtable}{lrrrrr}
\caption{Set of robots used for multi-embodiment training and design.}
\label{tab:robots_all} \\
\toprule
Robot & p-gain & d-gain & $\sigma_{\mathrm{robot}}$ & $n_u$ & $d_{\mathrm{design}}$ \\
\midrule
\endfirsthead
\toprule
Robot & p-gain & d-gain & $\sigma_{\mathrm{robot}}$ & $n_u$ & $d_{\mathrm{design}}$ \\
\midrule
\endhead
\midrule
\multicolumn{6}{r}{Continued on next page.} \\
\endfoot
\bottomrule
\endlastfoot

\multicolumn{6}{l}{\textit{Quadrupeds}} \\
Unitree A1          & $20.0$ & $0.5$ & $0.25$ & $12$ & $358$ \\
Unitree Go1         & $20.0$ & $0.5$ & $0.25$ & $12$ & $358$ \\
Unitree Go2         & $20.0$ & $0.5$ & $0.30$ & $12$ & $358$ \\
Unitree B2          & $80.0$ & $2.0$ & $0.50$ & $12$ & $358$ \\
ANYmal B            & $80.0$ & $2.0$ & $0.50$ & $12$ & $358$ \\
ANYmal C            & $80.0$ & $2.0$ & $0.50$ & $12$ & $358$ \\
Barkour v0          & $20.0$ & $0.5$ & $0.25$ & $12$ & $358$ \\
Barkour vB          & $30.0$ & $0.5$ & $0.25$ & $12$ & $358$ \\
Silver Badger       & $20.0$ & $0.5$ & $0.25$ & $13$ & $385$ \\
Honey Badger        & $20.0$ & $0.5$ & $0.25$ & $12$ & $358$ \\
Bittle              & $25.0$ & $0.5$ & $0.60$ & $8$  & $250$ \\
Aibo                & $20.0$ & $0.5$ & $0.25$ & $17$ & $493$ \\
Solo 12             & $20.0$ & $0.5$ & $0.25$ & $12$ & $358$ \\
Spot                & $80.0$ & $2.0$ & $0.50$ & $12$ & $358$ \\
Spot Mini           & $20.0$ & $0.5$ & $0.25$ & $12$ & $358$ \\

\midrule
\multicolumn{6}{l}{\textit{Bipeds and humanoids}} \\
Bolt                    & $15.0$  & $0.3$ & $0.20$ & $6$  & $190$ \\
Open Duck Mini          & $6.5$   & $0.5$ & $0.50$ & $16$ & $460$ \\
Mini Pi                 & $20.0$  & $0.5$ & $0.30$ & $12$ & $352$ \\
QMini                   & $15.0$  & $1.0$ & $0.40$ & $10$ & $298$ \\
Tron 1                  & $40.0$  & $1.0$ & $0.50$ & $6$  & $190$ \\
Unitree H1              & $60.0$  & $2.0$ & $0.75$ & $19$ & $541$ \\
Unitree G1              & $45.0$  & $1.0$ & $0.50$ & $23$ & $985$ \\
Talos                   & $80.0$  & $2.0$ & $0.75$ & $30$ & $1020$ \\
ROBOTIS OP3             & $21.0$  & $0.5$ & $0.60$ & $20$ & $568$ \\
NAO v5                  & $30.0$  & $1.0$ & $0.70$ & $22$ & $1102$ \\
ADAM Lite               & $80.0$  & $2.0$ & $0.75$ & $25$ & $703$ \\
AgiBot X1               & $45.0$  & $1.0$ & $0.70$ & $29$ & $811$ \\
Apptronik Apollo        & $100.0$ & $2.0$ & $0.75$ & $30$ & $890$ \\
Atlas                   & $80.0$  & $2.0$ & $0.75$ & $27$ & $783$ \\
Berkeley Humanoid       & $20.0$  & $0.6$ & $0.50$ & $12$ & $352$ \\
Berkeley Humanoid Lite  & $20.0$  & $0.5$ & $0.30$ & $22$ & $661$ \\
Booster T1              & $25.0$  & $0.6$ & $0.50$ & $23$ & $649$ \\
ELF2                    & $35.0$  & $0.7$ & $0.50$ & $25$ & $703$ \\
EngineAI SA01           & $55.0$  & $1.5$ & $0.50$ & $12$ & $352$ \\
Fourier GR1 T2          & $80.0$  & $2.0$ & $0.75$ & $34$ & $946$ \\
Fourier GR2 v3          & $80.0$  & $2.0$ & $0.75$ & $29$ & $811$ \\
Fourier N1              & $70.0$  & $2.0$ & $0.70$ & $23$ & $649$ \\
K-Bot v1                & $40.0$  & $2.0$ & $0.50$ & $20$ & $594$ \\
K-Bot v2                & $50.0$  & $1.0$ & $0.50$ & $20$ & $594$ \\
MIT Humanoid            & $30.0$  & $0.5$ & $0.50$ & $18$ & $514$ \\
Northwestern Humanoid   & $20.0$  & $0.6$ & $0.50$ & $10$ & $298$ \\
Poppy Humanoid          & $20.0$  & $0.5$ & $0.30$ & $25$ & $703$ \\
Sigmaban                & $21.0$  & $0.5$ & $0.50$ & $20$ & $568$ \\
STAR1                   & $80.0$  & $2.0$ & $0.75$ & $31$ & $1177$ \\
Valkyrie                & $100.0$ & $2.0$ & $0.75$ & $26$ & $730$ \\
Z-Bot                   & $10.0$  & $0.5$ & $0.40$ & $20$ & $594$ \\

\midrule
\multicolumn{6}{l}{\textit{Hexapods}} \\
Crab                  & $20.0$ & $0.5$ & $0.50$ & $18$ & $526$ \\
Custom Hexapod        & $30.0$ & $0.5$ & $0.60$ & $18$ & $526$ \\
Tom Hexapod           & $25.0$ & $0.5$ & $0.50$ & $18$ & $526$ \\
Golem                 & $25.0$ & $0.5$ & $0.50$ & $24$ & $688$ \\
\end{longtable}

\subsection{PPO Training}
\label{app:setup_rl}

All embodiment-aware policies and critics are trained with \gls{ppo}.
We use the hyperparameters in \autoref{tab:ppo_hyperparameters}.
Individual experiments deviate mainly in the number of training robots, and therefore in the number of parallel environments and total environment steps.

\begin{table}[h]
\centering
\def\arraystretch{1.2}
\caption{
\textbf{PPO hyperparameters.}
These are the default training hyperparameters. Large multi-robot runs adjust the total number of environment steps and number of parallel environments.
}
\label{tab:ppo_hyperparameters}
\small
\begin{tabular}{l l}
\toprule
Hyperparameter & Value \\
\midrule
Rollout length & $64$ \\
Minibatch size & $16{,}384$ \\
Epochs & $10$ \\
Learning rate schedule & linear, $1.5{\times}10^{-3} \rightarrow 3{\times}10^{-4}$ \\
Discount factor $\gamma$ & $0.99$ \\
GAE $\lambda$ & $0.7$ \\
Clip range & $0.1$ \\
Entropy coefficient & $0.005$ \\
Max gradient norm & $3.0$ \\
KL threshold & $0.5$ \\
Initial action standard deviation & $1.0$ \\
Policy mean clipping & $[-10,10]$ \\
Policy std. clipping & $[0.01,2.0]$ \\
Parallel environments & $4{,}096$ or $7{,}680$ or $15{,}040$ \\
Total environment steps & $1.0{\times}10^9$ or $3.7{\times}10^9$ \\
\bottomrule
\end{tabular}
\end{table}

Single-robot models are trained on randomized versions of one robot with $4{,}096$ parallel environments for $1.0{\times}10^9$ environment steps.
Morphology class models are trained on all robots of one morphology class except the held-out target robot, using $7{,}680$ parallel environments for $3.7{\times}10^9$ environment steps.
The all-robot model is trained on all $50$ robots with $15{,}040$ parallel environments for $3.7{\times}10^9$ environment steps.

Training was done on a mix of NVIDIA RTX 3090, A5000, A6000, RTX 6000 Ada, and V100 GPUs.
The largest experiment with all 50 robots took around $2.5$ days on a single RTX 6000 Ada.

\subsection{Design space parameters}
\label{app:design_space_table}

\autoref{tab:design_space} lists all parameter groups of the design space with their dimensions, transforms, and ranges.
The differentiable map $\Phi$ applies each normalized design parameter $f_i \in [-1,1]$ to its nominal value $x_{\mathrm{nom}}$ either multiplicatively as $x = x_{\mathrm{nom}} (1 + \kappa f_i)$ or additively as $x = x_{\mathrm{nom}} + \kappa f_i$, where $\kappa$ is the range of the parameter group.
The rotation groups rotate the nominal orientation by the angle $\kappa f_i$ around a fixed axis.
For a robot with $n_b$ bodies, $n_j$ joints, $n_u$ actuated joints, and $n_f$ feet, this leads to a total design dimensionality of $d_{\mathrm{design}} = 9 + 13 n_b + 11 n_j + 3 n_u + 3 n_f$.

Several consistency constraints inside $\Phi$ keep the designs physically valid.
The body size has one scaling factor for each spatial axis, which scales all body, joint, and geometry positions along that axis.
The masses are scaled with the cube of the mean of the three factors and the inertias with the fifth power of the factor of their axis.
The geometry sizes, torque and velocity limits, joint damping, armature, stiffness, friction loss, PD gains, and action scaling factor are scaled linearly with the mean of the three factors.
The coupled mass and inertia group scales the mass and the inertia of a body together, while the two decoupled groups change them individually.
Joint damping, armature, stiffness, and friction loss are clamped to be non-negative, and the nominal joint positions are clipped to the joint position limits.
For robots with closed kinematic loops, the body size, body position, and body orientation groups are disabled.

\begin{table}[h]
\centering
\caption{
\textbf{Parameter groups of the design space.}
We show the dimension, transform, and range $\kappa$ of each group for a robot with $n_b$ bodies, $n_j$ joints, $n_u$ actuated joints, and $n_f$ feet.
}
\label{tab:design_space}
\small
\begin{tabular}{lccc}
\toprule
Parameter group & Dimension & Transform & Range \\
\midrule
\multicolumn{4}{l}{\textit{Robot}} \\
Body size per axis         & $3$     & mult. & $\pm 80\%$ \\
IMU position               & $3$     & add.  & $\pm 0.1$\,m \\
\midrule
\multicolumn{4}{l}{\textit{Bodies}} \\
Coupled mass and inertia   & $n_b$   & mult. & $\pm 25\%$ \\
Decoupled mass             & $n_b$   & mult. & $\pm 25\%$ \\
Decoupled inertia          & $3 n_b$ & mult. & $\pm 25\%$ \\
Center of mass             & $3 n_b$ & add.  & $\pm 0.02$\,m \\
Inertia orientation        & $n_b$   & rot.  & $\pm 0.2$\,rad \\
Body position              & $3 n_b$ & add.  & $\pm 0.02$\,m \\
Body orientation           & $n_b$   & rot.  & $\pm 0.02$\,rad \\
\midrule
\multicolumn{4}{l}{\textit{Feet}} \\
Foot size                  & $3 n_f$ & mult. & $\pm 75\%$ \\
\midrule
\multicolumn{4}{l}{\textit{Joints}} \\
Joint axis                 & $n_j$   & rot.  & $\pm 0.1$\,rad \\
Joint range limits         & $2 n_j$ & add.  & $\pm 0.4$\,rad \\
Damping                    & $n_j$   & mult. & $\pm 100\%$ \\
Damping                    & $n_j$   & add.  & $\pm 0.1$\,N\,m\,s/rad \\
Armature                   & $n_j$   & mult. & $\pm 100\%$ \\
Armature                   & $n_j$   & add.  & $\pm 0.001$\,kg\,m$^2$ \\
Stiffness                  & $n_j$   & mult. & $\pm 10\%$ \\
Stiffness                  & $n_j$   & add.  & $\pm 0.1$\,N\,m/rad \\
Friction loss              & $n_j$   & mult. & $\pm 100\%$ \\
Friction loss              & $n_j$   & add.  & $\pm 10^{-5}$\,N\,m \\
\midrule
\multicolumn{4}{l}{\textit{Actuators}} \\
Torque limit               & $n_u$   & mult. & $\pm 75\%$ \\
Velocity limit             & $n_u$   & mult. & $\pm 75\%$ \\
Nominal joint position     & $n_u$   & add.  & $\pm 0.1$\,rad \\
\midrule
\multicolumn{4}{l}{\textit{Controller}} \\
P gain                     & $1$     & mult. & $\pm 50\%$ \\
D gain                     & $1$     & mult. & $\pm 50\%$ \\
Action scaling factor      & $1$     & mult. & $\pm 50\%$ \\
\bottomrule
\end{tabular}
\end{table}


\section{Direct-Design URMA Critic}
\label{app:critic_architecture}

\begin{figure}[!ht]
\centering
\includegraphics[width=\textwidth]{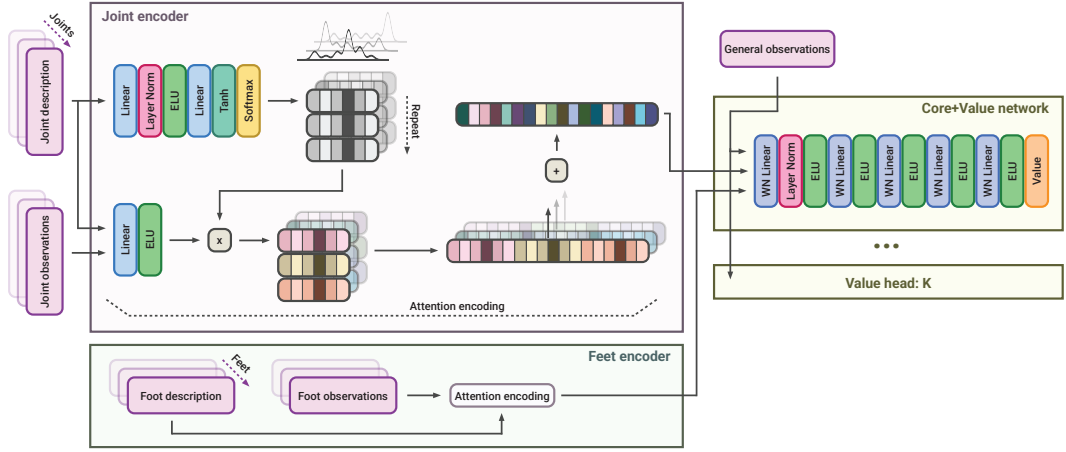}
\caption{
\textbf{Overview of the direct-design URMA critic architecture.}
We simplify the visualization: The foot encoder corresponds to the joint encoder, but with foot observations and foot description vectors as input, and the value heads all use the same structure.
The linear layers in the value heads are all wrapped in WeightNorm (WN) layers \cite{salimans2016weight}, which improves the conditioning of the critic optimization \cite{palenicek2026xqc}.
}
\label{fig:urma_critic_architecture}
\end{figure}

\autoref{fig:urma_critic_architecture} shows the direct-design critic used in our experiments.
Additionally, \autoref{tab:critic_vs_policy_arch} shows the layer widths and parameter counts.
Compared to the original \gls{urma} critic, the per-joint and per-foot encoders receive the corresponding description vectors in addition to the observations.
This gives the embodiment parameters a direct path into the value prediction.

\begin{table}[!ht]
\centering
\caption{Critic architecture ablation for gradient-based design on an early-stage Unitree Go2 design task, with results averaged over 10 initial designs.}
\label{tab:critic_codesign_ablation}
\begin{tabular}{lccc}
\hline
Critic & Heads & Latent dim. & $\Delta R$ \\
\hline
Original URMA critic & 1 & 4  & +10.39 \\
Original URMA critic & 3 & 4  & +13.57 \\
Original URMA critic & 3 & 8  & +12.59 \\
Original URMA critic & 3 & 16 & +10.22 \\
Direct-design critic  & 3 & 4  & +6.41  \\
Direct-design critic  & 3 & 8  & +9.12  \\
Direct-design critic  & 3 & 16 & +10.98 \\
Direct-design critic  & 3 & 32 & \textbf{+17.76} \\
Direct-design critic  & 3 & 64 & +14.35 \\
\hline
\end{tabular}
\end{table}

The ablations in \autoref{tab:critic_codesign_ablation} show that the direct-design architecture only becomes beneficial once the latent of the attention values has sufficient capacity.
With the default \gls{urma} latent dimension of $4$, the direct-design critic is bottlenecked and underperforms even the original critic.
Increasing the latent dimension improves the direct-design critic up to dimension $32$, where it gives the strongest design improvement across all variants.
Increasing the dimension further does not help, while increasing the original critic also does not close the gap.
We therefore use the direct-design \gls{urma} critic with $K=3$ heads and value-latent dimension $32$ throughout all our experiments.

\begin{table}[!ht]
\centering
\caption{Layer widths and parameter counts of policy and critic.}
\label{tab:critic_vs_policy_arch}
\begin{tabular}{lcc}
\hline
Block & Critic & Policy \\
\hline
Joint encoder         & 64, 64                & 256, 256 \\
Joint value latent    & 32                    & 4 \\
Foot encoder          & 32, 32                & --- \\
Foot value latent     & 32                    & --- \\
Head MLP              & 512, 256, 128, 128, 128, 1 & 512, 256, 256, 256, 256 \\
\# heads              & 3                     & 1 \\
\hline
Total parameters      & $5{,}378{,}312$        & $945{,}430$ \\
\hline
\end{tabular}
\end{table}


\section{Surrogate Optimizer Hyperparameters}
\label{app:optimizer_hyperparameters}

All optimizers use the same frozen direct-design \gls{urma} critic and optimize the same design optimization objective $\hat J_\lambda(f)$ from \autoref{eq:soft_trust_region}, with a minibatch size of $64$ states from the fixed state bank.
We set $\lambda=100$ for the trust-region penalty.
An ablation over $\lambda$ values is presented in \autoref{app:exp_trust_region_ablation}.
All final results are evaluated by rolling out the frozen policy $\pi_\theta$ on the found design $f^\star$ for 200 episodes and reporting the mean return.

We tuned the optimizer-specific hyperparameters on an early-stage version of the perturbed-start experiment on all three robots and made sure that all optimizers converged and used roughly similar budgets with respect to the number of iterations and critic evaluations.
Their hyperparameters are as follows:

\paragraph{Value-Gradient Design Search.}
\gls{vgds} optimizes \autoref{eq:soft_trust_region} with default Adam for $50$ iterations.
The learning rate is $0.05$ and each normalized design-parameter update is clipped to $\delta_{\max}=0.1$.

\paragraph{Random search.}
Random search evaluates $3200$ designs sampled uniformly from the normalized design space and returns the design with the highest value of $\hat J_\lambda(f)$.

\paragraph{Bayesian optimization.}
\gls{bo} uses the BoTorch SingleTaskGP with a Mat\'ern kernel and qLogNoisyExpectedImprovement as the acquisition function.
The optimizer starts from $10$ Sobol initial points, with the first point set to the initial design, and then runs for $50$ Bayesian optimization iterations.
The acquisition function is optimized with $8$ restarts, $256$ raw samples, and batch size $q=1$.

\paragraph{Covariance matrix adaptation evolution strategy.}
\gls{cmaes} is run for $50$ generations with initial step size $\sigma_0=0.5$ and population size
$4+\lfloor 3\ln(d_{\mathrm{design}})\rfloor$.
We use the diagonal-covariance parameterization due to the high dimensionality of the design space.

\paragraph{Trust-region Bayesian optimization.}
\gls{turbo} is implemented as adaptive trust-region random search.
It runs for $50$ iterations with $64$ candidate designs per iteration.
The trust-box radius is initialized to $0.1$, with minimum radius $0.005$ and maximum radius $0.5$.
The shrink and expansion factors are $0.7$ and $1.3$, respectively.

\paragraph{Particle swarm optimization.}
\gls{pso} uses the global-best variant with $32$ particles for $50$ iterations.
We use inertia $w=0.9$, cognitive coefficient $c_1=0.5$, and social coefficient $c_2=0.3$.

\paragraph{Cross-entropy method.}
\gls{cem} is run for $50$ iterations with $32$ particles.
Each iteration samples from a diagonal Gaussian, selects the top $20\%$ elites, and refits the mean and standard deviation to those elites.
The initial standard deviation is $0.5$, and the standard-deviation floor is $0.02$.

\paragraph{Differential evolution.}
\gls{de} is run for $50$ iterations with $32$ particles.
For each particle, we form a mutant vector from three randomly selected particles using differential weight $F=0.5$, apply binomial crossover with probability $CR=0.9$, and keep the trial candidate if it improves the objective.

\paragraph{Augmented random search.}
\gls{ars} uses antithetic finite differences with $16$ directions for $50$ iterations.
We use perturbation standard deviation $0.1$, learning rate $0.05$, and top-direction fraction $1.0$.

\paragraph{Gradient-covariance particle filter optimization.}
\gls{gcpfo} is run for $50$ iterations with $2$ regions and $32$ particles per region.
At each iteration, particles are resampled according to their objective values and updated with gradient-covariance-shaped exploration noise.
We use step size $0.2$, initial temperature $\tau_0=1.0$, and $\epsilon=10^{-8}$.


\section{Trust-Region and Reference-Design Ablations}
\label{app:exp_trust_region_ablation}

The value gradients are only useful if search stays in a region where the value function is reliable.
We therefore perform a sweep over the trust-region coefficient $\lambda$ on the single-robot design setting, where $\lambda=0$ effectively means no trust-region.
\autoref{tab:lambda_ablation} shows that without the trust-region penalty, performance drops a lot.
On the Unitree Go2, the final design is significantly worse than the initial random design, while the MIT Humanoid and Golem improve only slightly.
For $\lambda>0$, performance improves on all three robots.
Unitree Go2 performs best at $\lambda=100$.
MIT Humanoid and Golem reach their highest values at larger penalties, but the differences are small compared to the gap between $\lambda=0$ and any useful trust region.
We use $\lambda=100$ for the main experiments as it is the best value on Unitree Go2, close to the best values on the other robots, and avoids tuning the penalty separately per robot.
For the absolute best performance, tuning $\lambda$ per robot can be a good option for practical applications.
We additionally report a trust-region-only ablation (TR-only), where the backpropagation through the critic is removed and the optimizer only follows the penalty toward the reference design.
Under Adam, this update is invariant to the value of $\lambda$ which corresponds to the limit of $\lambda \rightarrow \infty$, therefore the design and its performance converges to the reference design based on the number of iterations $N$.

\begin{table}[h]
\centering
\caption{
\textbf{Trust-region sensitivity.}
Mean return improvement $\Delta R$ of \gls{vgds} on the single-robot design setting as a function of the trust-region penalty coefficient $\lambda$ over $10$ initial designs.
}
\label{tab:lambda_ablation}
\footnotesize
\setlength{\tabcolsep}{2.8pt}
\begin{tabular}{lccccccccccc}
\toprule
Robot & $\lambda=0$ & $\lambda=5$ & $\lambda=10$ & $\lambda=20$ & $\lambda=50$ & $\lambda=100$ & $\lambda=150$ & $\lambda=200$ & $\lambda=300$ & TR-only & $f_{\mathrm{ref}}$ \\
\midrule
Go2   & -18.93 & 14.51 & 14.30 & 14.18 & 13.93 & \textbf{20.60} & 14.18 & 14.12 & 13.87 & 19.60 & 19.81 \\
MIT   &   2.21 & 29.08 & 29.77 & 30.39 & 30.95 & 28.84 & 31.15 & \textbf{31.20} & 30.97 & 27.97 & 28.09 \\
Golem &   0.50 &  6.03 &  9.94 &  9.73 & 14.08 & 16.85 & 15.63 & 16.46 & \textbf{17.27} & 16.35 & 16.60 \\
\bottomrule
\end{tabular}
\end{table}

Next, we run \gls{vgds} directly from the reference design instead of a random initial design.
\autoref{tab:fromref_ablation} shows that as \gls{vgds} moves away from the reference it finds better designs in most settings.
Improving a new robot directly from its nominal design is a natural use-case of \gls{vgds} although the gain in performance is smaller than when starting from a random design, as the reference design is often an already good starting point in our setting.

\begin{table}[h]
\centering
\caption{
\textbf{Starting at the reference design.}
Return of the reference design $f_{\mathrm{ref}}$ and of the design found by \gls{vgds} when the search starts directly at $f_{\mathrm{ref}}$.
}
\label{tab:fromref_ablation}
\small
\setlength{\tabcolsep}{3pt}
\begin{tabular}{lcccccc}
\toprule
& \multicolumn{2}{c}{Go2} & \multicolumn{2}{c}{MIT} & \multicolumn{2}{c}{Golem} \\
\cmidrule(lr){2-3} \cmidrule(lr){4-5} \cmidrule(lr){6-7}
Training set & $f_{\mathrm{ref}}$ & \gls{vgds} & $f_{\mathrm{ref}}$ & \gls{vgds} & $f_{\mathrm{ref}}$ & \gls{vgds} \\
\midrule
Single robot                & $41.09$ & $\mathbf{42.04}$ & $30.34$ & $\mathbf{30.88}$ & $19.61$ & $\mathbf{20.21}$ \\
Morphology class (held out) & $37.83$ & $\mathbf{39.00}$ & $\mathbf{14.52}$ & $14.34$ & $1.27$  & $\mathbf{24.65}$ \\
All 50 robots               & $37.33$ & $\mathbf{39.18}$ & $17.40$ & $\mathbf{20.49}$ & $\mathbf{7.60}$  & $7.58$ \\
\bottomrule
\end{tabular}
\end{table}

Finally, we test the sensitivity to the quality of the reference design by using the random initial designs as the reference, i.e., $f'_{\mathrm{ref}} = f_{\mathrm{init}}$.
\autoref{tab:degraded_ref} shows that \gls{vgds} still improves performance even with the degraded reference designs.
The looser trust region with $\lambda=20$ works better than $\lambda=100$ here, as the search has to move further away from the worse anchor.

\begin{table}[h]
\centering
\caption{
\textbf{Degraded reference designs.}
Return of random reference designs $f'_{\mathrm{ref}} = f_{\mathrm{init}}$ and of the designs found by \gls{vgds} when the trust region is anchored at $f'_{\mathrm{ref}}$ instead of the nominal design.
We show the mean $\pm$ standard deviation over 3 initial designs in the single-robot design setting.
}
\label{tab:degraded_ref}
\small
\begin{tabular}{lccc}
\toprule
 & Go2 & MIT & Golem \\
\midrule
$f'_{\mathrm{ref}}$ & $27.59 \pm 4.27$ & $0.19 \pm 0.26$ & $2.33 \pm 3.22$ \\
\gls{vgds} with $\lambda=20$  & $\mathbf{37.16 \pm 1.69}$ & $\mathbf{18.49 \pm 11.41}$ & $\mathbf{7.00 \pm 4.95}$ \\
\gls{vgds} with $\lambda=100$ & $34.31 \pm 3.17$ & $16.90 \pm 10.98$ & $0.06 \pm 0.08$ \\
\bottomrule
\end{tabular}
\end{table}


\section{Critic Reliability Analysis}
\label{app:critic_reliability}

We analyze how reliable the frozen critic and its gradients remain as designs move away from the training distribution, and how strongly \gls{vgds} exploits critic errors during the search.
First, we evaluate designs $f = \alpha \cdot f^\star$ scaled along the found design direction, far beyond the designs that \gls{vgds} finds.
\autoref{tab:critic_linesearch} shows the rollout return, the mean value prediction, and the disagreement between the critic heads along the search direction.
Predicted value and rollout return stay aligned up to $\alpha \approx 2$--$5$, depending on the robot.
Also, the number of iterations $N$ was not sufficient to reach the optimum along the search direction, as both the return and value prediction still rise for $\alpha>1$.
Beyond this, the return starts to collapse while the critic still predicts high values, e.g., for the MIT Humanoid at $\alpha=5$.
Notably, the critic disagreement rises clearly before the collapse, which can provide an intrinsic warning signal.

\begin{table}[h]
\centering
\caption{
\textbf{Critic reliability along the found design direction.}
Rollout return $R$, mean value prediction $\bar V$, and critic head disagreement $\sigma_V$ for designs $f = \alpha \cdot f^\star$ in the single-robot design setting.}
\label{tab:critic_linesearch}
\small
\begin{tabular}{llcccccc}
\toprule
Robot & & $\alpha=0$ & $\alpha=1$ & $\alpha=2$ & $\alpha=3$ & $\alpha=5$ & $\alpha=8$ \\
\midrule
\multirow{3}{*}{Go2}   & $R$        & $41.0$ & $42.1$ & $42.4$ & $42.4$ & $38.6$ & $0.7$ \\
                       & $\bar V$   & $5.63$ & $5.80$ & $6.02$ & $5.98$ & $4.52$ & $1.60$ \\
                       & $\sigma_V$ & $0.04$ & $0.01$ & $0.12$ & $0.12$ & $0.61$ & $0.68$ \\
\midrule
\multirow{3}{*}{MIT}   & $R$        & $30.4$ & $30.7$ & $31.4$ & $30.6$ & $4.4$  & $0.0$ \\
                       & $\bar V$   & $5.32$ & $5.41$ & $5.46$ & $5.48$ & $5.03$ & $0.75$ \\
                       & $\sigma_V$ & $0.00$ & $0.00$ & $0.01$ & $0.02$ & $0.16$ & $0.41$ \\
\midrule
\multirow{3}{*}{Golem} & $R$        & $19.7$ & $19.8$ & $21.2$ & $9.6$  & $5.0$  & $0.0$ \\
                       & $\bar V$   & $4.96$ & $5.08$ & $5.20$ & $5.34$ & $2.20$ & $0.74$ \\
                       & $\sigma_V$ & $0.02$ & $0.01$ & $0.08$ & $0.23$ & $1.26$ & $0.46$ \\
\bottomrule
\end{tabular}
\end{table}

Next, we evaluate the value prediction and rollout return of the intermediate designs $f_t$ during the search, with and without the trust region.
\autoref{tab:critic_iterations} shows that the predicted value rises similarly in both settings.
With $\lambda=100$, the return rises together with the value prediction on all three robots.
With $\lambda=0$, the return instead collapses after $10$--$20$ iterations, while the predicted value continues to rise.
This highlights the critic exploitation that the trust region tries to prevent.

\begin{table}[h]
\centering
\caption{
\textbf{Critic exploitation over search iterations.}
Mean value prediction $\bar V$ from the start to the end of the search and rollout return $R$ of the intermediate designs $f_t$.
We show the mean over 3 initial designs in the single-robot design setting.
}
\label{tab:critic_iterations}
\small
\begin{tabular}{lccccccc}
\toprule
& $\bar V$ ($t{=}0 \to 50$) & \multicolumn{6}{c}{$R$ at iteration $t$} \\
\cmidrule(lr){3-8}
& & $0$ & $10$ & $20$ & $30$ & $40$ & $50$ \\
\midrule
Go2, $\lambda{=}100$   & $2.4 \to 5.8$ & $27.6$ & $40.3$ & $41.6$ & $41.4$ & $41.5$ & $41.8$ \\
Go2, $\lambda{=}0$     & $2.4 \to 6.2$ & $27.6$ & $25.5$ & $14.1$ & $12.3$ & $11.9$ & $8.7$ \\
MIT, $\lambda{=}100$   & $2.3 \to 5.4$ & $0.2$  & $27.8$ & $29.6$ & $30.8$ & $31.5$ & $31.4$ \\
MIT, $\lambda{=}0$     & $2.3 \to 5.6$ & $0.2$  & $10.1$ & $14.3$ & $7.9$  & $2.5$  & $2.4$ \\
Golem, $\lambda{=}100$ & $0.6 \to 5.2$ & $2.3$  & $11.3$ & $17.7$ & $17.8$ & $18.3$ & $20.0$ \\
Golem, $\lambda{=}0$   & $0.6 \to 5.5$ & $2.3$  & $4.7$  & $5.5$  & $5.4$  & $4.1$  & $2.8$ \\
\bottomrule
\end{tabular}
\end{table}

Finally, we sample random designs from the full design space and evaluate them scaled by factors of $0.25$--$1.0$, which moves the same design increasingly far away from the reference design.
Scaling a design by a factor $c$ is equivalent to sampling it from the smaller design space $[-c, c]^{d_{\mathrm{design}}}$.
\autoref{tab:critic_random} shows that both the return and value prediction degrade smoothly with the distance to reference.
The critic seems to be well calibrated in average over the growing design space that is induced by the curriculum and uniform sampling provides decent coverage of the per-dimension marginals around the reference design.
The critics overestimation that we observe during the design search seem to be a result of the exploitation of specific sub-spaces in the high dimensional design space rather than generic overestimation of the edges of the design space.

\begin{table}[h]
\centering
\caption{
\textbf{Critic reliability at increasing distance from the reference design.}
Rollout return $R$ and mean value prediction $\bar V$ of random designs from the full design space, scaled by a factor of $0.25$--$1.0$ toward the reference design.
We show the mean and $\pm$ standard deviation over 3 random designs per robot in the single-robot design setting.
}
\label{tab:critic_random}
\small
\begin{tabular}{llcccc}
\toprule
& & \multicolumn{4}{c}{Design scaling factor} \\
\cmidrule(lr){3-6}
Robot & & $0.25$ & $0.5$ & $0.75$ & $1.0$ \\
\midrule
\multirow{2}{*}{Go2}   & $R$        & $40.1 \pm 1.8$ & $36.7 \pm 4.1$  & $29.5 \pm 7.9$  & $19.0 \pm 13.1$ \\
                       & $\bar V$   & $5.60 \pm 0.08$ & $5.33 \pm 0.18$ & $4.13 \pm 0.47$ & $2.35 \pm 0.99$ \\
\midrule
\multirow{2}{*}{MIT}   & $R$        & $27.9 \pm 3.2$ & $13.1 \pm 11.1$ & $8.2 \pm 10.8$  & $5.0 \pm 7.0$ \\
                       & $\bar V$   & $5.32 \pm 0.01$ & $5.24 \pm 0.06$ & $4.47 \pm 0.52$ & $1.86 \pm 0.66$ \\
\midrule
\multirow{2}{*}{Golem} & $R$        & $18.7 \pm 2.3$ & $16.7 \pm 4.1$  & $12.5 \pm 3.4$  & $6.3 \pm 3.7$ \\
                       & $\bar V$   & $4.76 \pm 0.05$ & $3.96 \pm 0.23$ & $2.29 \pm 0.49$ & $1.00 \pm 0.05$ \\
\bottomrule
\end{tabular}
\end{table}

\section{Policy Fine-Tuning at Optimized Designs}
\label{app:policy_finetune}

To test whether further policy optimization helps after the design search, we fine-tune the frozen policy for 52M steps ($\approx 5\%$ of the pre-training budget) at the fixed optimized design $f^\star$ and, as a control test, at the initial design $f_{\mathrm{init}}$.
\autoref{tab:policy_finetune} shows that fine-tuning at $f^\star$ adds nothing on the Go2, where the frozen policy is already optimal, but adds $+3.4$ return on the MIT Humanoid and $+18.8$ on Golem.
Evaluating the frozen policy can therefore substantially underestimate the optimized designs.
Fine-tuning at $f_{\mathrm{init}}$ never closes the gap to the optimized designs and even destabilizes the policy on Golem.
We note that fine-tuning is a cheap way to squeeze out additional performance when the design is not already well covered by the generalist policy.

\begin{table}[h]
\centering
\caption{
\textbf{Policy fine-tuning at fixed designs.}
Return of the frozen policy and of the policy fine-tuned for 52M steps at the fixed initial and optimized designs.
We show the mean and $\pm$ standard deviation over 3 seeds for the fine-tuned policies in the single-robot design setting.
}
\label{tab:policy_finetune}
\small
\begin{tabular}{lcccc}
\toprule
& \multicolumn{2}{c}{$f_{\mathrm{init}}$} & \multicolumn{2}{c}{$f^\star$} \\
\cmidrule(lr){2-3} \cmidrule(lr){4-5}
Robot & Frozen & Fine-tuned & Frozen & Fine-tuned \\
\midrule
Go2   & $22.94$ & $28.47 \pm 0.24$ & $\mathbf{41.99}$ & $41.87 \pm 0.08$ \\
MIT   & $5.79$  & $5.74 \pm 0.13$  & $31.11$ & $\mathbf{34.49 \pm 0.17}$ \\
Golem & $12.43$ & $0.00 \pm 0.00$  & $20.34$ & $\mathbf{39.10 \pm 0.21}$ \\
\bottomrule
\end{tabular}
\end{table}

\section{Class-Balanced Training Distribution}
\label{app:balanced_training}

The all-robot training set is dominated by 15 quadrupeds and 31 bipeds and humanoids, with only 4 hexapods.
To isolate the effect of this class imbalance, we train an additional model on a class-balanced set of 12 robots (4 per morphology class, including the three target robots) with the same per-robot step budget as the all-robot model, and rerun the design experiments with the same 10 initial designs.
\autoref{tab:balanced_training} shows that the balanced critic improves the results on Golem, while Go2 and the MIT Humanoid decrease, as they can no longer draw on the many robots of their classes.
The hexapod-only critic stays ahead on Golem, as its class training set also has a ${\sim}4.5\times$ larger per-robot step budget and no other morphology classes to fit.
The found designs again outperform the reference design under the balanced critic on all three robots ($+0.7\,/\,+1.1\,/\,+1.4$).
In general, broad training sets seem to help well-represented classes, while underrepresented classes benefit from balanced or even more from specialized training sets.

\begin{table}[h]
\centering
\caption{
\textbf{Class-balanced training distribution.}
Return improvement $\Delta R$ of \gls{vgds} with critics trained on the target robot's morphology class, on the class-balanced 12-robot set, and on all 50 robots.
The morphology class critics hold the target robot out, while the other two training sets include it.
We show the mean and $\pm$ standard deviation over the same 10 initial designs.
}
\label{tab:balanced_training}
\small
\begin{tabular}{lccc}
\toprule
Robot & Morphology class (held out) & Balanced 12 & All 50 robots \\
\midrule
Go2   & $24.10 \pm 10.90$ & $23.33 \pm 9.78$ & $\mathbf{25.62 \pm 10.94}$ \\
MIT   & $13.75 \pm 1.09$  & $15.03 \pm 0.88$ & $\mathbf{20.47 \pm 1.30}$ \\
Golem & $\mathbf{13.65 \pm 9.33}$  & $6.79 \pm 3.03$  & $5.23 \pm 3.44$ \\
\bottomrule
\end{tabular}
\end{table}

\section{Sim-to-Real Transfer of the Training Pipeline}
\label{app:sim2real}

\begin{figure}[h]
\centering
\includegraphics[width=0.75\textwidth]{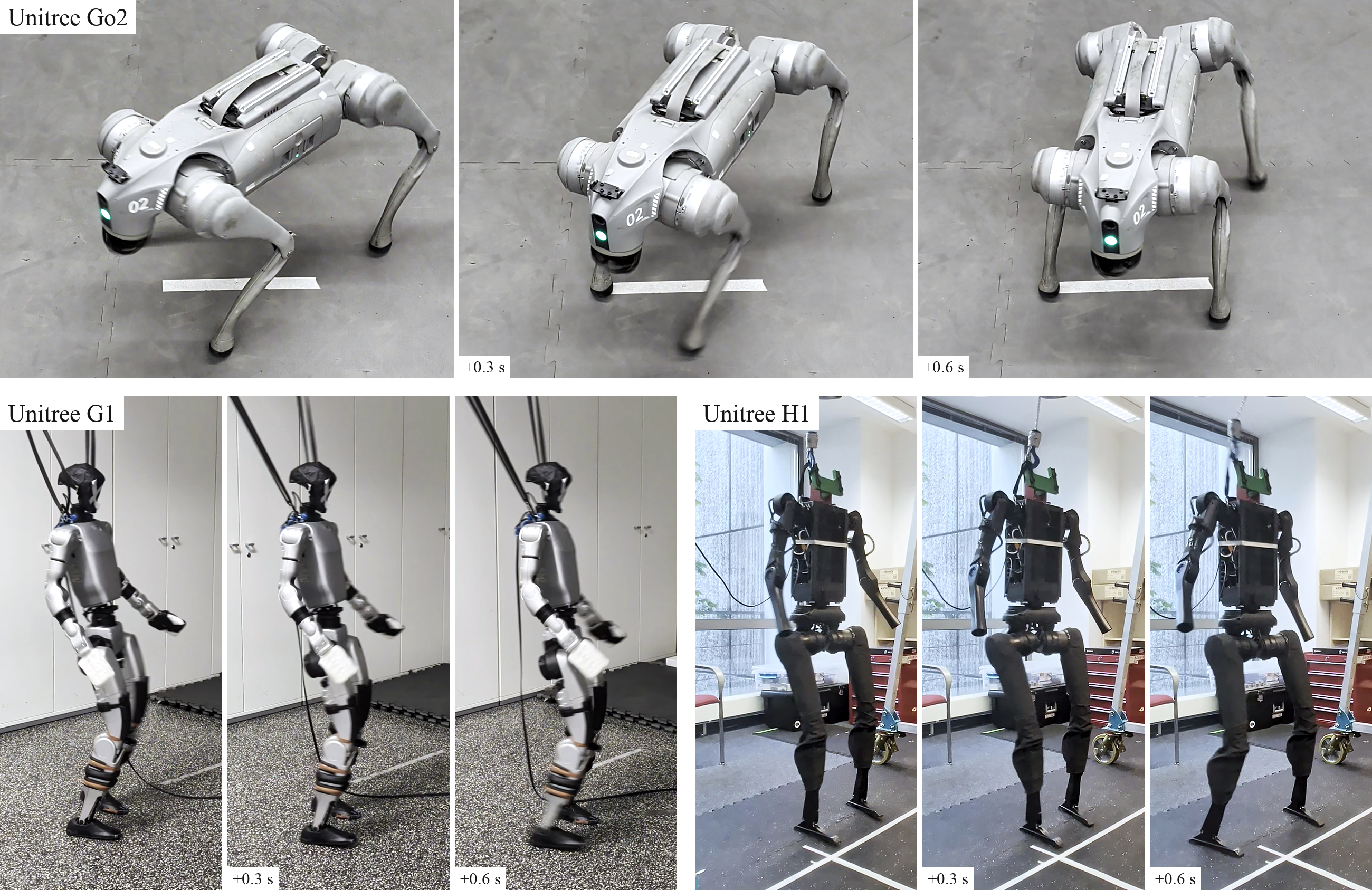}
\caption{
\textbf{Real-world deployment of the multi-embodiment policy at the nominal designs.}
We test typical omnidirectional walking on the Unitree Go2, Unitree G1, and Unitree H1.
}
\label{fig:sim2real_photos}
\end{figure}

While manufacturing the optimized designs is beyond the scope of this work, the training pipeline itself is sim-to-real transferable.
\autoref{fig:sim2real_photos} shows the deployment of the policy trained on all 50 robots on the real Unitree Go2, Unitree G1, and Unitree H1 at their nominal designs.

\section{Velocity Tracking Error Reduction Results}
\label{app:velocity_tracking_results}

As a more intuitive and task-specific metric, we also report the reduction in absolute $x$-$y$ linear velocity tracking error,
\begin{equation}
\Delta e_{xy} = e_{xy}(f_{\mathrm{init}}) - e_{xy}(f^\star),
\end{equation}
where positive values indicate lower tracking error after the design search.
Improvements in return, as shown in \autoref{fig:e1_return} and \autoref{fig:e2_e4_return}, do not always imply lower $x$-$y$ tracking error, especially when the optimized design trades tracking performance against other reward terms.

\begin{figure}[h]
\centering
\includegraphics[width=\textwidth]{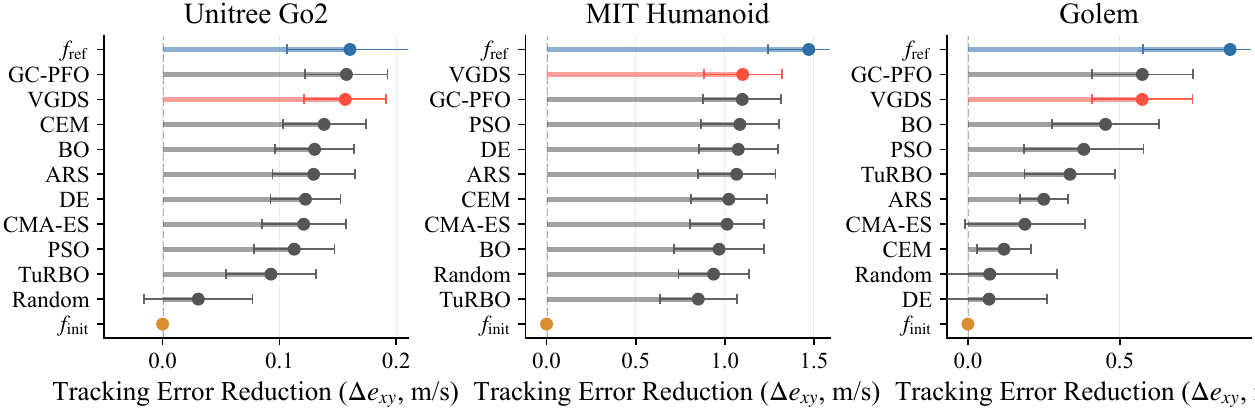}
\caption{
\textbf{$x$-$y$ tracking error reduction for single-robot design.}
Mean reduction $\Delta e_{xy}$ in absolute $x$-$y$ linear velocity tracking error relative to the initial perturbed design $f_{\mathrm{init}}$, for $10$ starts per robot.
}
\label{fig:e1_xy}
\end{figure}

\begin{figure}[h]
\centering
\includegraphics[width=\textwidth]{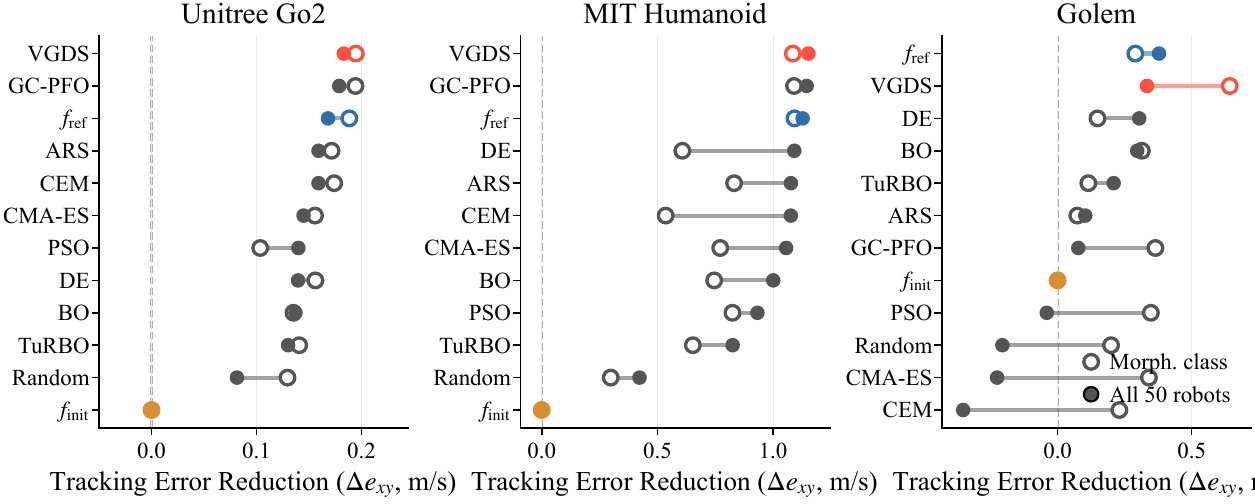}
\caption{
\textbf{$x$-$y$ tracking error reduction for multi-robot design.}
Mean reduction $\Delta e_{xy}$ in absolute $x$-$y$ linear velocity tracking error when the target robot is either held out from its morphology class training set or included in the full 50-robot training set.
We omit error bars for readability and report them in \autoref{app:numeric_results}.
}
\label{fig:e2_e4_xy}
\end{figure}

In the single-robot setting, \autoref{fig:e1_xy} closely follows the return results.
\gls{vgds} and \gls{gcpfo} give almost identical tracking error reductions on all three robots and are the strongest search methods overall.
For the training set comparison in \autoref{fig:e2_e4_xy}, the tracking metric shows the same broad trend for the three robots.

\section{Numeric Results for the Training Set Comparison}
\label{app:numeric_results}

\autoref{tab:numeric_e2_e4} and \autoref{tab:numeric_e2_e4_xy} report the numeric results of \autoref{fig:e2_e4_return} and \autoref{fig:e2_e4_xy} with their standard deviations.

\begin{table}[h]
\centering
\caption{
\textbf{Numeric results for the training set comparison.}
We show the mean $\pm$ standard deviation of the return improvement $\Delta R$ over the $10$ initial designs per robot, when the target robot is held out from its morphology class training set (Class) or included in the full 50-robot training set (All 50).
The last row shows the return of the initial designs.
}
\label{tab:numeric_e2_e4}
\footnotesize
\setlength{\tabcolsep}{2pt}
\begin{tabular}{@{}lcccccc@{}}
\toprule
& \multicolumn{2}{c}{Go2} & \multicolumn{2}{c}{MIT} & \multicolumn{2}{c}{Golem} \\
\cmidrule(lr){2-3} \cmidrule(lr){4-5} \cmidrule(lr){6-7}
Method & Class & All 50 & Class & All 50 & Class & All 50 \\
\midrule
\gls{vgds}     & $24.10 \pm 10.90$ & $\mathbf{25.62 \pm 10.94}$ & $13.75 \pm 1.09$ & $\mathbf{20.47 \pm 1.30}$ & $\mathbf{13.65 \pm 9.33}$ & $\mathbf{5.23 \pm 3.44}$ \\
Random         & $9.90 \pm 13.72$ & $8.54 \pm 13.22$ & $1.18 \pm 3.60$ & $1.88 \pm 3.37$ & $0.32 \pm 1.46$ & $-0.31 \pm 3.61$ \\
\gls{bo}       & $11.47 \pm 16.57$ & $15.04 \pm 13.57$ & $6.61 \pm 6.28$ & $8.65 \pm 5.90$ & $1.00 \pm 0.77$ & $-0.04 \pm 4.02$ \\
\gls{cmaes}    & $16.03 \pm 8.52$ & $16.84 \pm 12.77$ & $2.17 \pm 3.76$ & $7.67 \pm 4.08$ & $1.54 \pm 1.96$ & $0.01 \pm 3.95$ \\
\gls{turbo}    & $13.02 \pm 14.43$ & $15.43 \pm 15.16$ & $1.60 \pm 3.17$ & $6.47 \pm 5.80$ & $1.70 \pm 2.35$ & $-0.11 \pm 4.73$ \\
\gls{pso}      & $11.88 \pm 14.37$ & $14.50 \pm 8.13$ & $3.80 \pm 4.21$ & $10.18 \pm 7.35$ & $1.45 \pm 2.23$ & $0.40 \pm 4.55$ \\
\gls{cem}      & $20.51 \pm 10.22$ & $20.24 \pm 9.07$ & $4.47 \pm 4.66$ & $12.70 \pm 6.12$ & $1.62 \pm 1.62$ & $0.52 \pm 2.98$ \\
\gls{de}       & $16.39 \pm 9.47$ & $15.25 \pm 12.07$ & $0.88 \pm 2.85$ & $12.89 \pm 7.03$ & $0.62 \pm 0.80$ & $2.21 \pm 3.57$ \\
\gls{ars}      & $20.56 \pm 10.51$ & $21.81 \pm 10.96$ & $2.51 \pm 4.18$ & $10.40 \pm 5.79$ & $0.65 \pm 1.02$ & $1.07 \pm 4.13$ \\
\gls{gcpfo}    & $\mathbf{24.42 \pm 10.67}$ & $25.01 \pm 10.65$ & $\mathbf{13.82 \pm 2.25}$ & $18.11 \pm 3.38$ & $2.22 \pm 1.53$ & $1.65 \pm 3.34$ \\
\midrule
$f_{\mathrm{ref}}$ & $23.37 \pm 10.81$ & $23.63 \pm 10.69$ & $14.20 \pm 0.81$ & $17.04 \pm 0.96$ & $0.69 \pm 0.62$ & $5.13 \pm 3.46$ \\
\midrule
$R(f_{\mathrm{init}})$ & $14.46 \pm 10.80$ & $13.70 \pm 10.69$ & $0.32 \pm 0.82$ & $0.36 \pm 0.96$ & $0.58 \pm 0.62$ & $2.47 \pm 3.38$ \\
\bottomrule
\end{tabular}
\end{table}

\begin{table}[h]
\centering
\caption{
\textbf{Numeric tracking error results for the training set comparison.}
We show the mean $\pm$ standard deviation of the reduction $\Delta e_{xy}$ in absolute $x$-$y$ linear velocity tracking error in m/s over the $10$ initial designs per robot, when the target robot is held out from its morphology class training set (Class) or included in the full 50-robot training set (All 50).
The last row shows the tracking error of the initial designs.
}
\label{tab:numeric_e2_e4_xy}
\footnotesize
\setlength{\tabcolsep}{2pt}
\begin{tabular}{@{}lcccccc@{}}
\toprule
& \multicolumn{2}{c}{Go2} & \multicolumn{2}{c}{MIT} & \multicolumn{2}{c}{Golem} \\
\cmidrule(lr){2-3} \cmidrule(lr){4-5} \cmidrule(lr){6-7}
Method & Class & All 50 & Class & All 50 & Class & All 50 \\
\midrule
\gls{vgds}     & $\mathbf{0.19 \pm 0.14}$ & $\mathbf{0.18 \pm 0.11}$ & $\mathbf{1.09 \pm 0.58}$ & $\mathbf{1.15 \pm 0.59}$ & $\mathbf{0.64 \pm 0.50}$ & $\mathbf{0.33 \pm 0.50}$ \\
Random         & $0.13 \pm 0.13$ & $0.08 \pm 0.07$ & $0.30 \pm 0.53$ & $0.42 \pm 1.01$ & $0.20 \pm 0.64$ & $-0.21 \pm 0.90$ \\
\gls{bo}       & $0.14 \pm 0.17$ & $0.14 \pm 0.11$ & $0.75 \pm 0.50$ & $1.00 \pm 0.66$ & $0.31 \pm 0.50$ & $0.30 \pm 0.51$ \\
\gls{cmaes}    & $0.16 \pm 0.14$ & $0.14 \pm 0.11$ & $0.77 \pm 0.66$ & $1.06 \pm 0.59$ & $0.34 \pm 0.51$ & $-0.23 \pm 0.36$ \\
\gls{turbo}    & $0.14 \pm 0.16$ & $0.13 \pm 0.13$ & $0.65 \pm 0.74$ & $0.83 \pm 0.65$ & $0.11 \pm 0.84$ & $0.21 \pm 0.29$ \\
\gls{pso}      & $0.10 \pm 0.20$ & $0.14 \pm 0.10$ & $0.82 \pm 0.56$ & $0.93 \pm 0.63$ & $0.35 \pm 0.51$ & $-0.04 \pm 0.93$ \\
\gls{cem}      & $0.17 \pm 0.14$ & $0.16 \pm 0.11$ & $0.54 \pm 0.71$ & $1.08 \pm 0.60$ & $0.23 \pm 0.68$ & $-0.35 \pm 0.62$ \\
\gls{de}       & $0.16 \pm 0.14$ & $0.14 \pm 0.11$ & $0.61 \pm 0.63$ & $1.09 \pm 0.56$ & $0.15 \pm 0.23$ & $0.30 \pm 0.50$ \\
\gls{ars}      & $0.17 \pm 0.15$ & $0.16 \pm 0.12$ & $0.83 \pm 0.56$ & $1.08 \pm 0.56$ & $0.07 \pm 0.55$ & $0.10 \pm 0.29$ \\
\gls{gcpfo}    & $\mathbf{0.19 \pm 0.14}$ & $\mathbf{0.18 \pm 0.11}$ & $\mathbf{1.09 \pm 0.58}$ & $1.14 \pm 0.58$ & $0.36 \pm 0.52$ & $0.08 \pm 0.50$ \\
\midrule
$f_{\mathrm{ref}}$ & $0.19 \pm 0.14$ & $0.17 \pm 0.11$ & $1.09 \pm 0.58$ & $1.13 \pm 0.59$ & $0.29 \pm 0.49$ & $0.38 \pm 0.50$ \\
\midrule
$e_{xy}(f_{\mathrm{init}})$ & $0.30 \pm 0.14$ & $0.28 \pm 0.11$ & $1.37 \pm 0.58$ & $1.36 \pm 0.59$ & $0.77 \pm 0.49$ & $0.76 \pm 0.50$ \\
\bottomrule
\end{tabular}
\end{table}


\section{RL Co-Design Baselines}
\label{app:rl_baseline_adaptations}

We adapt Schaff2019~\cite{schaff2019jointly}, \gls{feacrl}, Transform2Act, BodyGen, and Stackelberg \gls{ppo} to the same high-dimensional locomotion co-design setting used in our main experiments.
The original methods were proposed for different robot representations, lower dimensional design spaces, or other base \gls{rl} algorithms.
The Transform2Act family of methods was originally proposed to also handle discrete topology optimization.
In our setting, we only evaluate the continuous-design variant, because our goal is to optimize fixed-topology robot models around plausible initial designs rather than to evolve new robots from scratch.
To make all the \gls{rl} methods compatible, we replace their normal policy and critic networks with \gls{urma}-style networks, and train them all in the on-policy setting with \gls{ppo}.
While these adaptations make the comparison controlled, they do not fully reproduce the original methods, and therefore the results should be interpreted as comparisons of the underlying co-design ideas rather than of the specific implementations.

\paragraph{Schaff2019.}
Schaff2019 maintains a distribution over designs and updates this distribution from rollout returns while training a design-conditioned policy.
We keep this population-based structure, but represent designs in the normalized embodiment space $f$ and train the policy with our \gls{ppo} implementation.
Each parallel environment samples a design from the current design distribution, and the distribution is shifted toward designs that achieve higher episode return.

\paragraph{Fast Evolutionary Actor-Critic Reinforcement Learning.}
\gls{feacrl} alternates between controller learning and design selection, using an actor-critic value estimate to score candidate designs.
The original method uses off-policy \gls{sac} and a learned state-action value function.
In our on-policy implementation, we replace \gls{sac} with \gls{ppo} and use the learned value function over a buffer of start states as the design surrogate.
Candidate designs are selected with \gls{pso} in the normalized design space.

\paragraph{Transform2Act.}
Transform2Act learns a transform policy that modifies the robot design before an execution policy controls the resulting body.
Our robots have fixed topology and continuous design parameters, so we use the continuous attribute variant and remove the discrete skeleton actions such as adding or removing joints.
At the beginning of each episode, the transform policy outputs a continuous design update in the normalized design space and the resulting embodiment is then used for the rollout of the execution policy.
Both the transform and execution policies are trained with \gls{ppo}.

\paragraph{BodyGen.}
BodyGen builds on the structure of Transform2Act and improves the architecture and credit assignment between design and control.
Because our robot topologies are fixed and the morphology is already represented by \gls{urma} description vectors, we do not use BodyGen's additional topology-aware self-attention.
However, we implement the stage-specific credit assignment.
In our setting, BodyGen therefore differs from Transform2Act mainly through its separate design and control value functions and the corresponding stage-specific advantage estimates.

\paragraph{Stackelberg PPO.}
Stackelberg \gls{ppo} builds on BodyGen and treats morphology optimization and control learning as a leader-follower problem, where the design policy acts as the leader and the control policy adapts as the follower.
We adapt this idea to the same fixed-topology continuous design space as the other baselines.
The leader proposes design updates in the normalized embodiment space, while the follower is a \gls{urma}-style \gls{ppo} policy trained on the resulting designs.


\section{All 50 Robots Design Results}
\label{app:full_robot_design_results}

\begin{figure}[t]
\centering
\includegraphics[width=\textwidth]{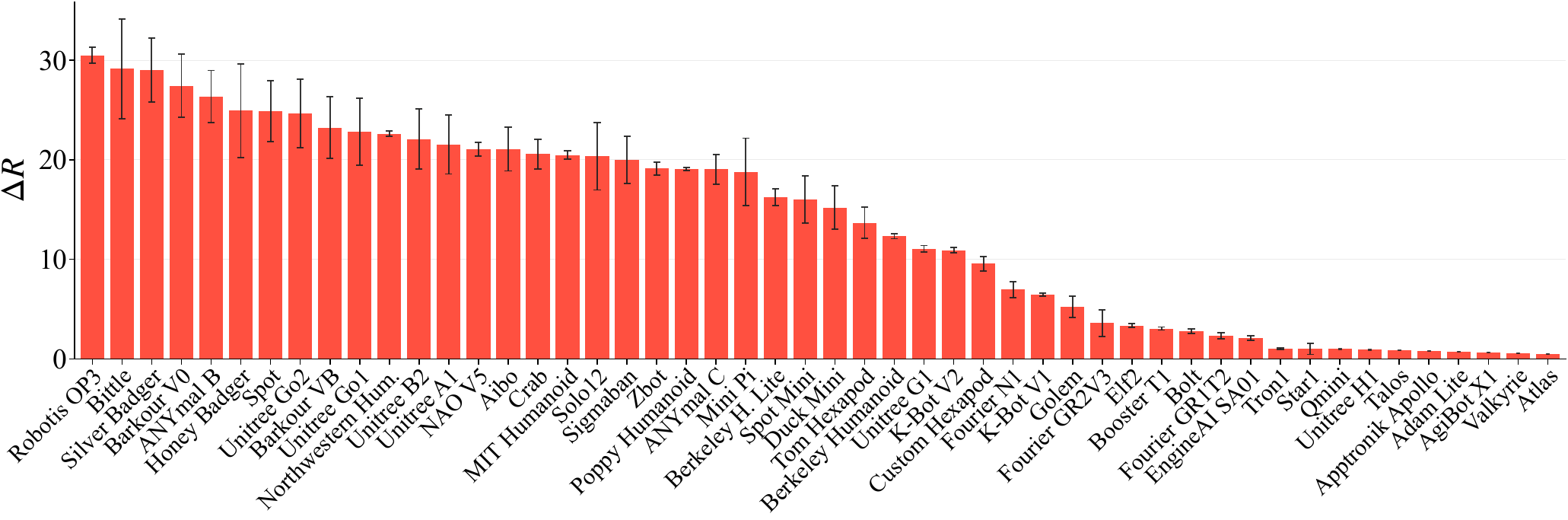}
\caption{
\textbf{Design search on all 50 training robots.}
Per-robot return improvement $\Delta R = R_{\mathrm{found}} - R_{\mathrm{init}}$ of \gls{vgds} applied to each robot in the full 50-robot training set, starting from $10$ uniformly sampled random initial designs per robot.
}
\label{fig:full_robot_results}
\end{figure}

We use the same trained policy and critic from the main experiments, which were trained on all 50 robots, to run \gls{vgds} on every robot in the training set, and report the return improvements in \autoref{fig:full_robot_results}.
We note that some of the large humanoid runs do not appear to have fully converged yet and would benefit from more training steps, which is the reason for the small improvements on some of those robots.


\section{Design Analysis Details}
\label{app:design_analysis_details}

We provide additional visualizations for the design analysis in \autoref{sec:exp_design_analysis}.
We compare optimized designs to the nominal reference in the normalized design space, $\Delta f = f^\star - f_{\mathrm{ref}}$, and group dimensions by body part and parameter type.
Each group is summarized by the \gls{rms} update magnitude and its dominant signed direction across design searches.

\begin{figure}[t]
\centering
\includegraphics[width=\textwidth]{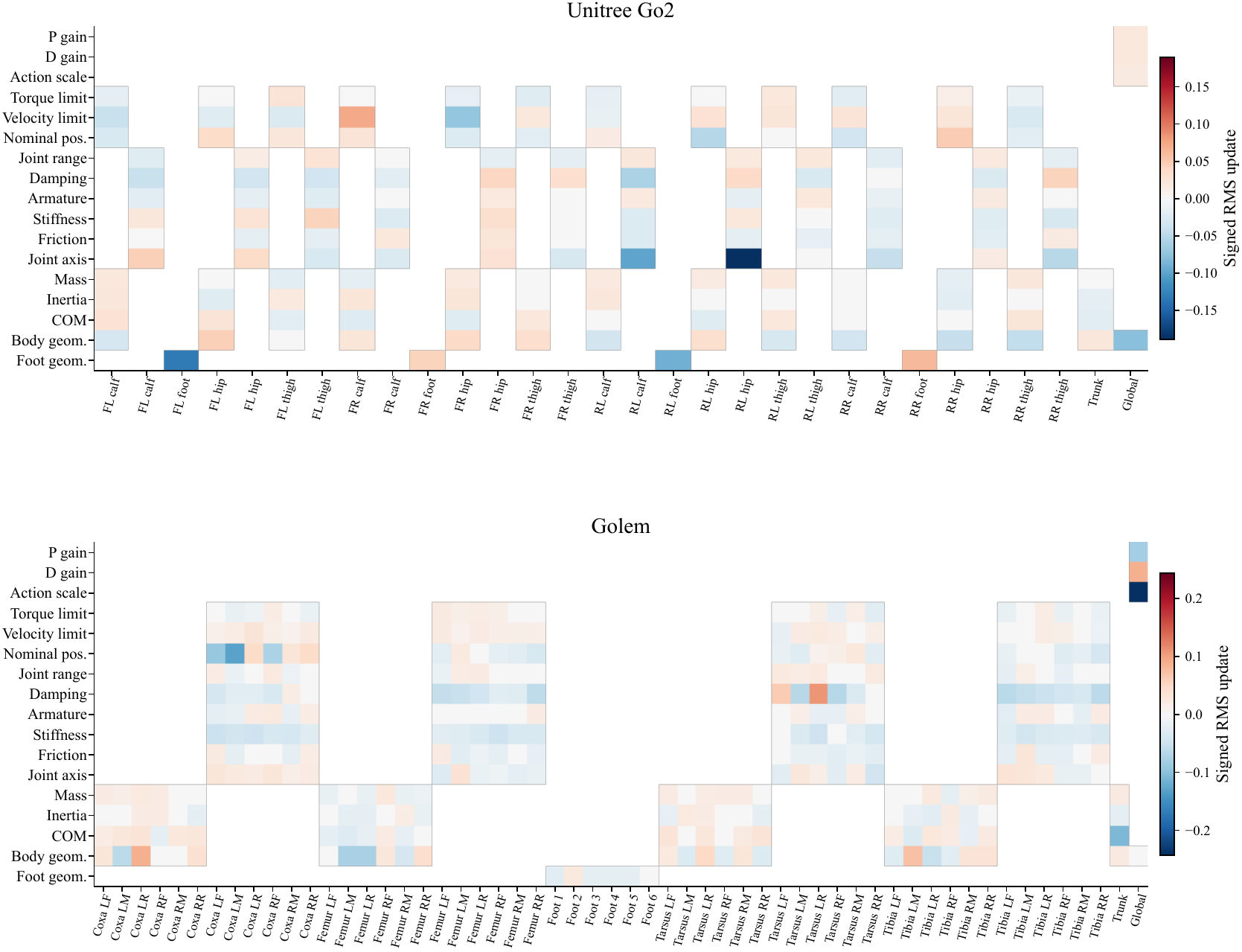}
\caption{
\textbf{Grouped design changes for the Unitree Go2 and Golem.}
Signed \gls{rms} changes of the optimized designs relative to the nominal reference, grouped by body part and parameter type.
The heatmaps show which design groups are changed most consistently by \gls{vgds}.
}
\label{fig:design_analysis_heatmaps}
\vspace{-1.0em}
\end{figure}

\autoref{fig:design_analysis_heatmap_mit} and \autoref{fig:design_analysis_heatmaps} show the full grouped changes.
MIT Humanoid has strong changes in nominal joint positions and foot geometry.
Unitree Go2 shows more localized changes in leg axes, foot geometry, and actuator limits.
Golem shows large control parameter changes, especially action scale and gains, together with nominal position changes in the coxa.

\begin{figure}[t]
\centering
\includegraphics[width=\textwidth]{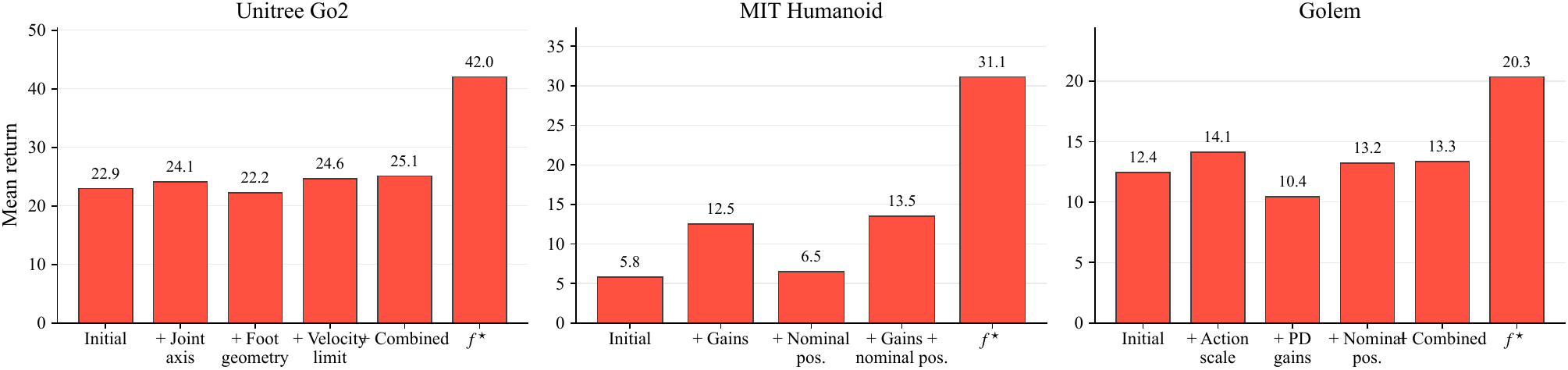}
\caption{
\textbf{Parameter group evaluations.}
Selected parameter groups from the optimized design $f^\star$ are copied into the initial design $f_{\mathrm{init}}$ and evaluated with the same policy.
}
\label{fig:design_analysis_counterfactuals}
\end{figure}

\autoref{fig:design_analysis_counterfactuals} evaluates selected groups in isolation and in combination, by substituting them into the initial design and evaluating the return improvement.
For MIT Humanoid, the optimized gains are the most useful isolated group.
For Unitree Go2, the best result comes from combining the highlighted joint axis, foot geometry, and actuator velocity limit changes.
For Golem, reducing the action scale helps, while changing the PD gains alone hurts performance.
In general, the partially modified designs recover part of the full improvement, but the complete optimized design remains substantially better.

As an additional sanity check, we verify that the optimized designs do not simply saturate the design bounds.
Across the three single-robot experiments, no optimized dimension exceeds a near-boundary threshold of $|f_i|>0.9$, and the average normalized change per dimension is small, between $0.027$ and $0.042$.
The largest systematic shift is the Golem's control parameters, where \gls{vgds} reduces the action scale by $12.2\%$, lowers the P gain by $4.1\%$, and increases the D gain by $4.4\%$.
On the $17$ evaluated partially modified designs, the value predictions are positively correlated with true rollout return, with Spearman $\rho=0.40$ overall and $\rho=0.50$, $0.60$, and $0.83$ for MIT Humanoid, Unitree Go2, and Golem.
The final design $f^\star$ has both the highest value prediction and the highest rollout return for all three robots.

Some optimized designs appear stretched or visually disconnected in our renderings.
This is an artifact of the visual meshes, which are not rescaled together with the underlying geometry, while the collision geometry and the kinematic tree actually remain intact.
We count the self-collisions of the collision geometry in the nominal pose for the $30$ optimized designs of the single-robot experiments.
All Unitree Go2 and MIT Humanoid designs have no self-collisions, and the Golem designs have up to $8$, compared to none for the reference design and up to $30$ for the initial designs.
The main failure case we observe is on some of the large humanoids in the 50-robot training set, where the policy has not fully converged and the design improvements remain near zero, as discussed in \autoref{app:full_robot_design_results}.


\section{Design Search Trajectories}
\label{app:design_trajectory_gallery}

\begin{figure}[t]
\centering
\includegraphics[width=\textwidth]{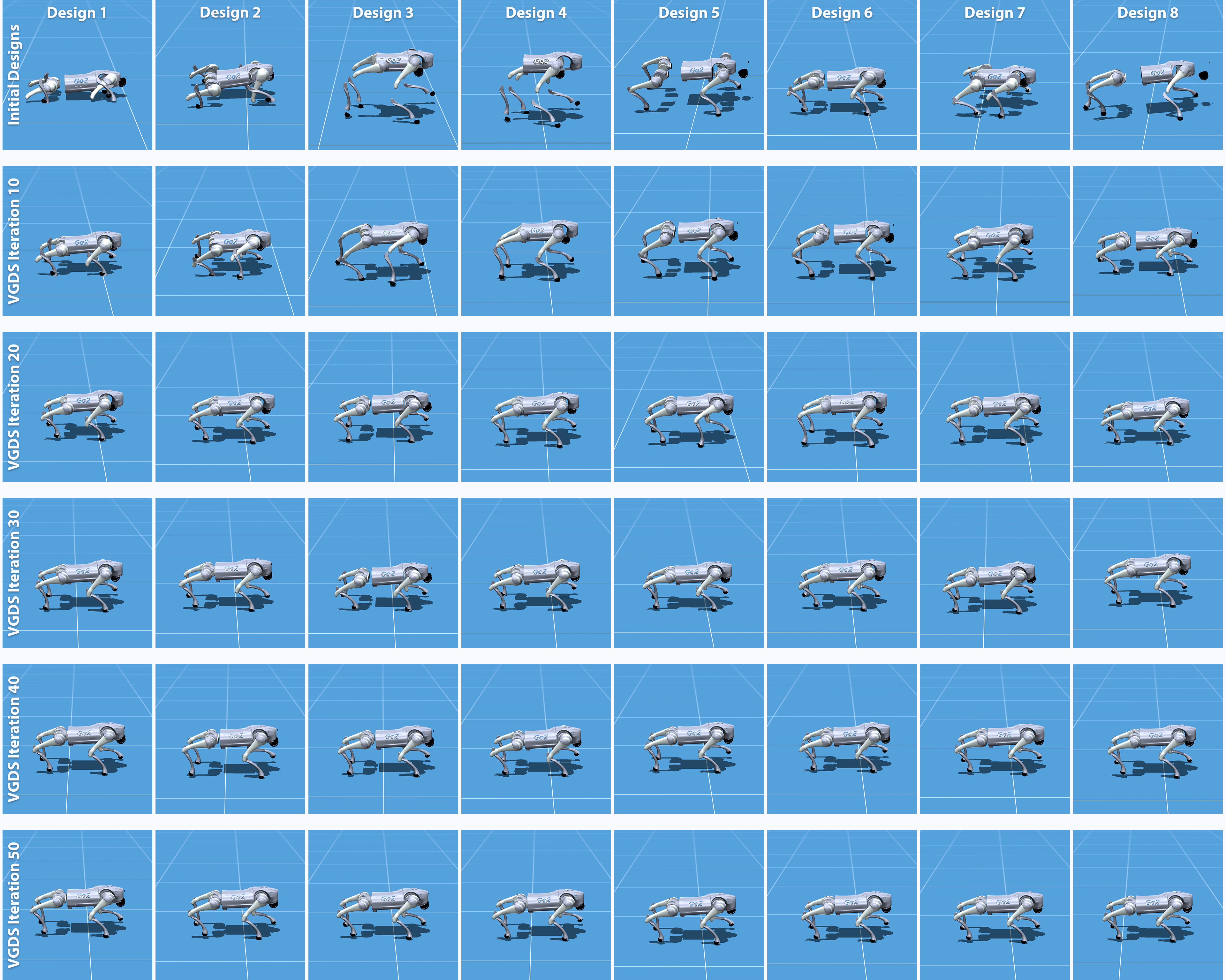}
\caption{
\textbf{Design search trajectories for the Unitree Go2 quadruped.}
The columns show 8 different initial designs, and the rows show the initial design and \gls{vgds} at iteration 10, 20, 30, 40, and 50.
}
\label{fig:go2_trajectory_gallery}
\end{figure}

\begin{figure}[t]
\centering
\includegraphics[width=\textwidth]{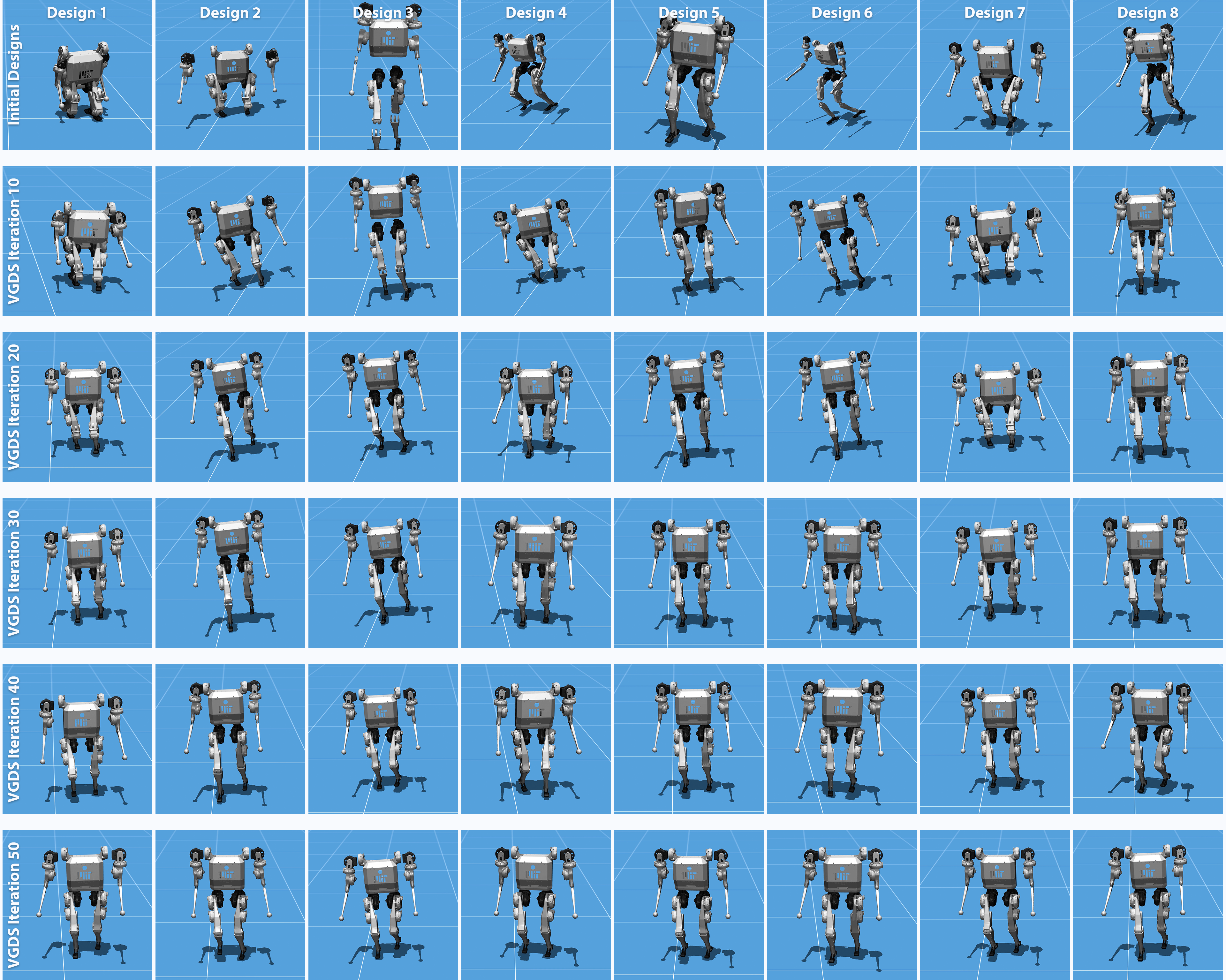}
\caption{
\textbf{Design search trajectories for the MIT Humanoid.}
The columns show 8 different initial designs, and the rows show the initial design and \gls{vgds} at iteration 10, 20, 30, 40, and 50.
}
\label{fig:mith_trajectory_gallery}
\end{figure}

\begin{figure}[t]
\centering
\includegraphics[width=\textwidth]{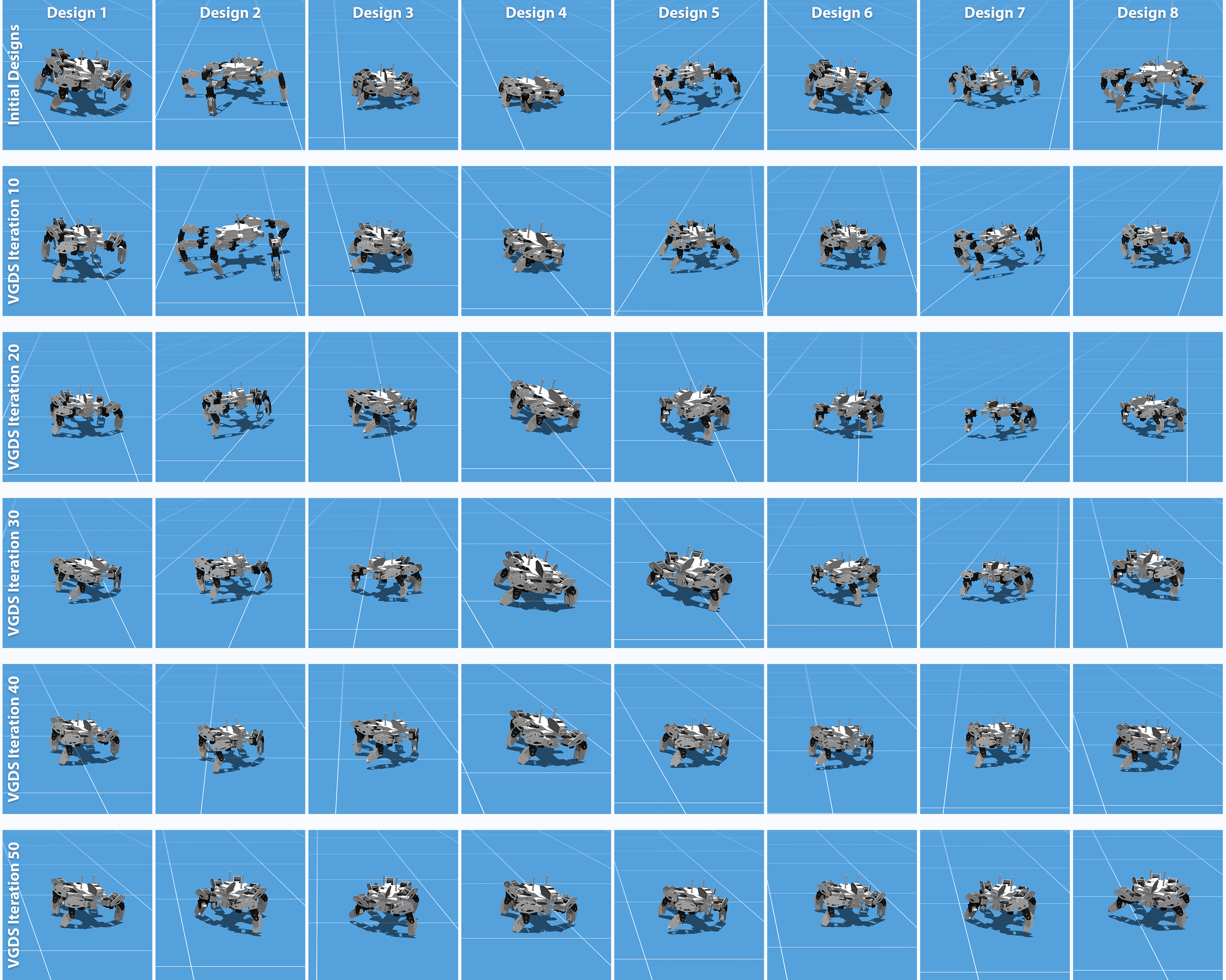}
\caption{
\textbf{Design search trajectories for the Golem hexapod.}
The columns show 8 different initial designs, and the rows show the initial design and \gls{vgds} at iteration 10, 20, 30, 40, and 50.
}
\label{fig:golem_trajectory_gallery}
\end{figure}

\autoref{fig:go2_trajectory_gallery}, \autoref{fig:mith_trajectory_gallery}, and \autoref{fig:golem_trajectory_gallery} show the \gls{vgds} optimization trajectories in the single-robot setting for 8 of the 10 initial designs used for the main experiments with the Unitree Go2, MIT Humanoid, and Golem, respectively.
While the initial designs all visually appear quite different, \gls{vgds} finds consistent improvements across all of them, and the designs all converge to a similar robot-specific optimum.
This shows a sort of diversity collapse, which is expected but not always welcome in design optimization.
Fewer iterations, a weaker trust region or a reduced learning rate can help to preserve more diversity, while trading off some of the final performance.

\clearpage

\clearpage

\clearpage


\section{Additional Method Variants}
\label{app:additional_method_variants}

During the development of \gls{vgds}, we tested several variants for the search, and while they did not outperform the final method, they provided useful insights into the design choices and failure modes.
We hope that reporting these negative results can help future work, so we summarize the most informative variants here.

\paragraph{Hard trust-region projection.}
We tested a hard trust region, where the design is projected back into a fixed-radius ball around the reference design after each Adam step.
This proved to be competitive with the soft L2 objective in most settings.
However, it introduces an additional radius that must be chosen explicitly, and we found that tuning this radius was slightly more difficult than tuning the smoother L2 penalty coefficient $\lambda$.

\paragraph{Uncertainty-based trust regions.}
A natural alternative is to use disagreement between critic heads as a trust region.
We tried absolute and relative uncertainty formulations, as well as lower confidence bound objectives.
While these versions could get rid of the dependence on a reference design, they did not provide a reliable trust region in the end.
We found that the small ensemble of value heads, trained on the same data, could still agree on some catastrophic extrapolations.
We still believe such an uncertainty-based trust region is a promising direction and \autoref{app:critic_reliability} shows that there is a signal in the critic disagreement, which needs to be integrated and calibrated properly to be a useful trust region.

\paragraph{Search dynamics.}
We tested several optimizer modifications, including initial design noise, per-step gradient noise, Adam moment resets, L2-norm update clipping, cosine learning rate schedules, SGD instead of Adam, and sweeps over learning rate, step size, and number of iterations.
None of these changes produced consistent improvements over the simple Adam configuration that we ended up using.
More search iterations can sometimes even hurt when over-optimizing the critic.

\paragraph{Multi-start search.}
We also tested a multi-start version, where multiple particles are initialized with small perturbations, moved through the gradients, and the best particle is selected by the surrogate objective.
We found that once the trust region objective is chosen well, multi-start search did not lead to consistent improvements over the simpler single-design version.

\paragraph{Direction and subspace structure.}
We tested whether the useful design update is concentrated in a small number of dimensions.
Active subspace variants that restricted updates to the largest gradient dimensions degraded performance.
Similarly, removing one dominant search direction was mostly harmless, but restricting the update to only that direction collapsed the performance.
This showed us that the useful design signal is mostly dense and high-dimensional.

\paragraph{Refreshing the state bank.}
One natural concern of the fixed state bank idea is that it becomes inconsistent as the design changes during search.
We therefore tried refreshing the states by rolling out the policy under the current design during optimization or sampling states from the starting-state distribution of the current design.
This did not improve results and was on rare occasions even worse.
We hypothesize that the fixed state bank works because the generalist policy is robust enough that the state distributions across closely related embodiments heavily overlap, and the trust region keeps the new design within that valid neighborhood.


\end{document}